\documentclass[10pt,twocolumn,letterpaper]{article}

\usepackage{cvpr}              
\usepackage{graphicx}
\usepackage{caption}
\usepackage{stfloats}
\usepackage{enumitem}
\usepackage{cuted}
\usepackage{tocloft}
\usepackage{titletoc}
\usepackage{overpic}
\usepackage{capt-of}  
\makeatletter
\let\oldmaketitle\maketitle
\renewcommand{\maketitle}{
    \oldmaketitle

    \begin{strip}
        \centering
        \vspace{-10mm}
        \includegraphics[width=\textwidth]{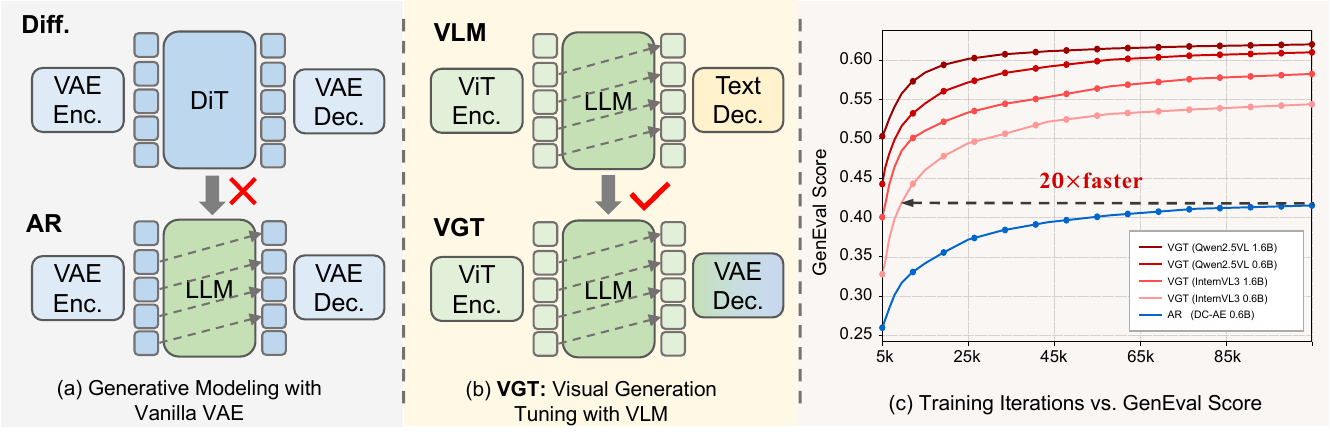}
        \vspace{-2mm}

        \captionof{figure}{
            Comparisons of autoregressive generation architectures.
            (a) Prior continuous autoregressive models directly regress VAE latents produced by pixel-level diffusion VAEs.
            (b) Our VGT pre-aligns the pretrained VLM semantic encoder with the latent space of the pixel decoder, eliciting emergent generation capabilities.
            (c) VGT achieves up to \textbf{20×} faster training than the VAE-based autoregressive generation (DC-AE) baseline. 
        }
        \label{fig:teaser}
        \vspace{-5pt}
    \end{strip}
}
\makeatother

\definecolor{cvprblue}{rgb}{0.21,0.49,0.74}
\usepackage[pagebackref,breaklinks,colorlinks,allcolors=cvprblue]{hyperref}
\usepackage{colortbl} 
\usepackage{xcolor}    
\usepackage{amssymb} 
\usepackage{multirow} 
\usepackage{pifont}
\usepackage{pifont}
\usepackage{marvosym}  
\usepackage[table]{xcolor}

\def\paperID{536} 
\def\confName{CVPR}
\def\confYear{2026}

\def\methodname{VGT}

\title{Visual Generation Tuning}

\author{
Jiahao Guo\textsuperscript{$\rm 1,4^{* \ddagger}$},
Sinan Du\textsuperscript{$\rm 2,4^{* \ddagger}$},
Jingfeng Yao\textsuperscript{$\rm 1$},
Wenyu Liu\textsuperscript{$\rm 1$},
Bo Li\textsuperscript{$\rm 4$},
Haoxiang Cao\textsuperscript{$\rm 3,4^\ddagger$} \\
Kun Gai\textsuperscript{\rm 4},
Chun Yuan\textsuperscript{\rm 2},
Kai Wu\textsuperscript{$\rm 4^{\dagger}$},
Xinggang Wang\textsuperscript{\rm 1 \Letter}\\
[1ex]
\normalsize \textsuperscript{\rm 1}Huazhong University of Science and Technology (HUST),
\textsuperscript{\rm 2}Tsinghua University \\
\normalsize\textsuperscript{\rm 3}School of Artificial Intelligence, South China Normal University,
\textsuperscript{\rm 4}Kolors Team, Kuaishou Technology
}
\begin{document}
\maketitle

\let\thefootnote\relax\footnotetext{$*$ Equal Contibution. $\dagger$ Project Lead. \Letter Corresponding Authors.}
\let\thefootnote\relax\footnotetext{$\ddagger$ Work done during internship in Kolors Team, Kuaishou Technology.}

\begin{abstract}
Large Vision Language Models (VLMs) effectively bridge the modality gap through extensive pretraining, acquiring sophisticated visual representations aligned with language. 
However, it remains underexplored whether these representations, optimized for multimodal understanding tasks, harbor an inherent potential for visual generation. 
In this paper, we propose \textbf{VGT}, \underline{\textbf{V}}isual \underline{\textbf{G}}eneration \underline{\textbf{T}}uning, a novel paradigm designed to stimulate the underlying capabilities of visual generation within \textbf{any} vision language models. 
By performing efficient visual generation tuning on well-pretrained VLMs, we significantly mitigate the alignment costs and accelerate the convergence of autoregressive modeling in the continuous space(\textbf{20×} speedup). 
Specifically, we dismiss the entangled pixel-level VAEs designed for diffusion transformers and formulate \textbf{VGT-AE} through aligning the semantic encoders from pretrained VLMs with the latent representations of pixel decoders. 
In image reconstruction tasks, we achieve \textbf{26.67} PSNR and \textbf{0.50} rFID at a \textbf{28×} compression ratio, outperforming specialized VAEs; In visual generation tasks, we achieve state-of-the-art outcomes among autoregressive models, \textbf{0.77} on GenEval and \textbf{78.73} on DPG-Bench. 
Furthermore, our proposed VGT showcases significant scaling promise and is versatile for endowing any VLMs trained for multimodal understanding with the capabilities of visual generation, which paves the new avenue to explore next-generation unified multimodal foundation models.Models and codes are available at \url{https://github.com/hustvl/VGT}.
\end{abstract} 
\vspace{-18pt}

\section{Introduction}
\label{sec:intro}

Autoregressive modeling~\cite{llamagen,intern-s1,internvl3,Qwen2_5_VL,gpt4,gemini} has emerged as a dominant paradigm in language and multimodal generation, demonstrating promising potential for application in visual generation. Previous autoregressive visual generation models primarily relied on vector quantization~\cite{vqvae,vqgan,vqkd} codebooks to convert images into discrete tokens, inherently introducing quantization errors. MAR~\cite{mar} addresses this through a lightweight flow matching head designed to predict the continuous latent representations of VAEs~\cite{kl-16}.

However, existing methods~\cite{mar,fluid,nextstep_1, spherear} neglect a fundamental dilemma: the latent representations learned by vanilla VAEs are poorly aligned with autoregressive modeling. Since VAEs are optimized for pixel-level reconstruction, their latent representations lack semantic structure, which leads to training instability and frequent variance collapse~\cite{sun2024multimodal,nextstep_1, spherear}. Inspired by recent studies~\cite{rae,svg,unilip} highlighting that structured semantic representations (\textit{e.g.}, CLIP~\cite{clip} and DINOv2~\cite{dino}) can improve the stability and efficiency in training diffusion transformers (DiTs), we explore a critical question: Vision language models (VLMs) are inherently well-aligned with discriminative visual representations; \textit{\textbf{Could we directly leverage this property to transfer visual generation capabilities into well-pretrained VLMs, thereby achieving unified models capable of both multimodal understanding and visual generation?}}

In this paper, we propose \textbf{VGT}, \underline{\textbf{V}}isual \underline{\textbf{G}}eneration \underline{\textbf{T}}uning, a novel paradigm designed to elicit emergent capabilities of visual generation inherently in \textbf{\textit{any}} VLMs originally trained for multimodal understanding tasks. 
VGT substantially lowers the costs associated with alignment while simultaneously promoting faster convergence in continuous-space autoregressive modeling. 
Specifically, we formulate \textbf{VGT-AE}, which builds upon the semantic encoder of well-pretrained VLMs and trains a pixel decoder for fine-grained reconstruction using a two-stage training strategy. 
In the first stage, the semantic structure of latent features is preserved via self-distillation loss. 
Subsequently, channel normalization and noise regularization~\cite{sun2024multimodal,nextstep_1} are applied to enhance the robustness of the latent representations and their adaptability for generative tuning.
Inspired by order-agnostic autoregressive methods~\cite{mar,fluid,rar,randar} that facilitate global modeling, we introduce a position-query mechanism that maintains the autoregressive formulation during training while enabling partially parallel decoding during inference. 
VGT is highly versatile and compatible with various VLMs, \textit{e.g.}, InternVL3~\cite{internvl3}. 
Moreover, as illustrated in Figure~\ref{fig:teaser}c, our VGT demonstrates significant data efficiency compared to vanilla VAE-based autoregressive models. 
Our contributions are summarized as follows:
\begin{enumerate}
    \item We propose a novel visual generation tuning (\textbf{VGT}) paradigm that enables the generation ability inherent in any VLM originally trained on multimodal understanding tasks.
    \item We propose \textbf{VGT-AE}, which compresses the semantic features from the VLM vision encoder into a compact structured latent representation for high-fidelity reconstruction and continuous AR modeling.
    \item Extensive experiments demonstrate the versatility of our VGT paradigm, and we achieve SOTA performance in reconstruction \textbf{(26.67 PSNR and 0.50 rFID)} and generation \textbf{(0.77 GenEval and 78.73 DPG-Bench)} with significant data efficiency \textbf{(20x speedup)}.
\end{enumerate}

\begin{figure*}[ht]
    \centering
    \includegraphics[width=1.0\linewidth]{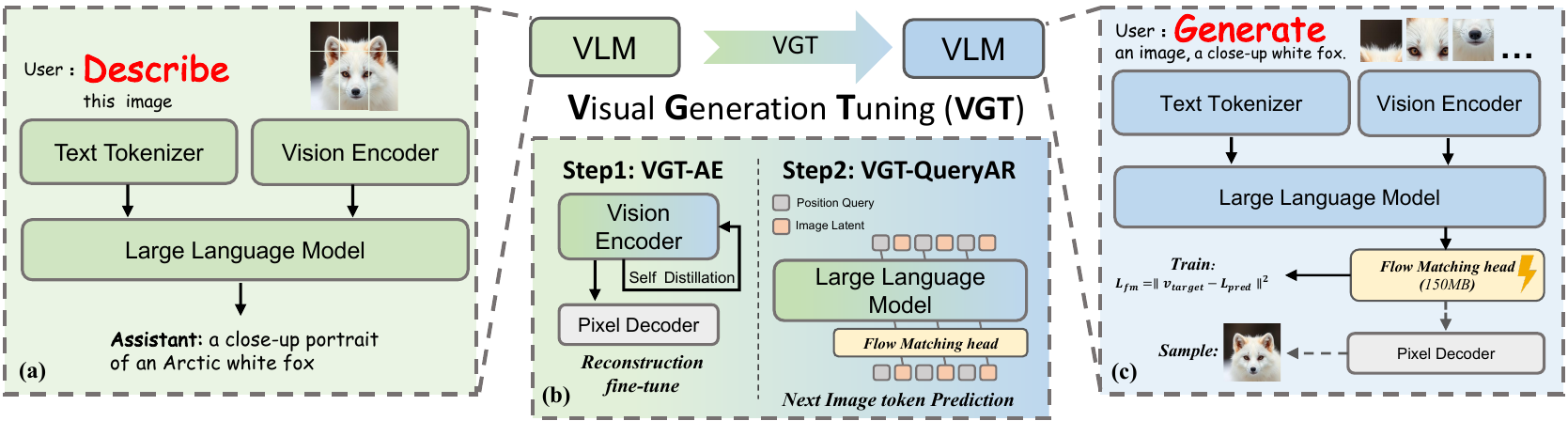}
    \caption{
    \textbf{An overview of our {\methodname} paradigm, which transforms a pre-trained Vision-Language Model (VLM) into a powerful generative model.} (a) Pre-trained VLMs excel at visual understanding by aligning semantic vision encoders with Language Models (LLMs), hinting at their latent image generation capabilities. (b) Our {\methodname}-AE (Visual Generation Tuning-AutoEncoder) aligns the VLM's semantic encoder through reconstruction training. (c) We perform visual generation tuning({\methodname}) on the VLM by predicting these semantically aligned image latents using a lightweight flow matching head, enabling efficient continuous autoregressive image generation.
    }
    \label{fig:method}
    \vspace{-8pt}
\end{figure*}

\section{Related Work}
\label{sec:related}

\subsection{Visual Tokenizers for Generative Modeling}

\textbf{Discrete.} Early autoregressive visual generation methods predominantly employed vector quantization (VQ) techniques to convert images into discrete token sequences. VQ-VAE~\cite{vqvae} pioneered this approach, with subsequent improvements including VQGAN~\cite{vqgan}, which incorporated adversarial and perceptual losses. Follow-up works enhanced codebook utilization and semantic alignment through hierarchical quantization~\cite{vqgan-lc}, multi-codebook schemes~\cite{simvq}, and bit-efficient representations~\cite{ibq}. Methods such as VQKD~\cite{vqkd}, TokenFlow~\cite{tokenflow}, and UniTok~\cite{unitok} further integrated semantic understanding into tokenizer training, though they face inherent trade-offs between reconstruction fidelity and semantic comprehension.



\hspace{-0.4cm}\textbf{Continuous.} In diffusion models, variational autoencoders (VAEs)~\cite{vae} serve as the primary framework for learning continuous latent representations. Enhanced variants including KL-16~\cite{kl-16}, FluxVAE~\cite{fluxvae}, and VA-VAE~\cite{vavae, vavae_dit} optimize the balance between compression efficiency and reconstruction quality. Recent approaches have explored leveraging pretrained visual representations: REPA~\cite{repa} and REPA-E~\cite{repae} use semantic features from semantic encoder to guide the denoising process and improve VAE representations, while RAE~\cite{rae}, ClipGen~\cite{clipgen}, and SVG~\cite{svg} directly employ semantically rich pretrained features for generation. However, these methods predominantly operate within diffusion frameworks using high-dimensional representations, making them unsuitable for efficient integration with autoregressive modeling.
We bridge this gap by leveraging semantic encoders from well-pretrained VLMs to align with low-dimensional latent representations of VAEs for continuous autoregressive generation. 

\subsection{Autoregressive visual generation}


\textbf{Next-Scale Prediction} methods such as VAR~\cite{var} employ a coarse-to-fine generation strategy across multiple resolution scales, utilizing bidirectional context modeling within each scale to capture both local details and global structure. \textbf{Next-Set Prediction} methods, commonly implemented as masked generative modeling, include MaskGIT~\cite{maskgit} and MAR~\cite{mar}. These approaches predict subsets of tokens in parallel using bidirectional attention mechanisms, achieving favorable trade-offs between generation efficiency and sample quality through iterative refinement processes. \textbf{Next-token Prediction} methods follow the causal autoregressive paradigm of language models, generating images token by token in sequential order. Early approaches like VQGAN~\cite{vqgan} and LlamaGen~\cite{llamagen}, used discrete tokens with raster-scan ordering, while recent methods including Ming-UniVision~\cite{ming-univision}, and NextStep-1~\cite{nextstep_1} have explored continuous token representations. Within this paradigm, researchers have investigated various design choices including discrete versus continuous token representations~\cite{givt,sun2024multimodal} and fixed raster order versus randomized generation orders~\cite{randar,rar}. However, these continuous autoregressive methods predominantly borrow the VAEs designed for pixel-level diffusion models, negelecting the misalignment with autoregressive modeling.


\section{Method}
\label{sec:method}

\begin{table*}[h!]
\centering
\scalebox{0.98}{
\begin{tabular}{cc|cc|cccc|cc}
\toprule
\multirow{2}{*}{\textbf{\# Exp.}} & \multirow{2}{*}{\textbf{Self-}} & \multicolumn{2}{c|}{\textbf{Reconstruction}} & \multicolumn{4}{c|}{\textbf{Understanding}} & \multicolumn{2}{c}{\textbf{Generation}} \\
\cmidrule(lr){3-4} \cmidrule(lr){5-8} \cmidrule(lr){9-10}
& \textbf{Distillation} & \textbf{rFID$\downarrow$} & \textbf{PSNR$\uparrow$} & \textbf{MME-P$\uparrow$} & \textbf{MMB$\uparrow$} & \textbf{AI2D$\uparrow$} & \textbf{TQA$\uparrow$} & \textbf{GenEval$\uparrow$} & \textbf{DPG-Bench$\uparrow$} \\
\midrule
1 & \ding{55} & \textbf{0.66} & \textbf{26.26} &  550.8 & 24.8 & 45.4 & 4.4 & 0.43 &  65.67 \\
2 & \ding{51} & 1.13 & 23.95 & \textbf{1495.6} & \textbf{72.6} & \textbf{69.6} & \textbf{73.9} & \textbf{0.49} & \textbf{66.90} \\
\rowcolor{cyan!10}
\bottomrule
\end{tabular}
}
\caption{
\textbf{Pilot experiments that indicate the potential trade-off between understanding, reconstruction and generation.}
We observe that over-optimizing the autoencoder towards reconstruction leads to a degradation in generative capability, while employing a self-distillation loss to constrain the latent representations to preserve semantic structures can enhance robustness in generative modeling. Previous works validate this insight under the settings of DiTs while we further demonstrate it in the continuous AR paradigm.
\vspace{-10pt}
}
\label{tab:key_findings}
\end{table*}

\subsection{Motivation}
\begin{figure}[t]
\centering
\includegraphics[width=1\linewidth]{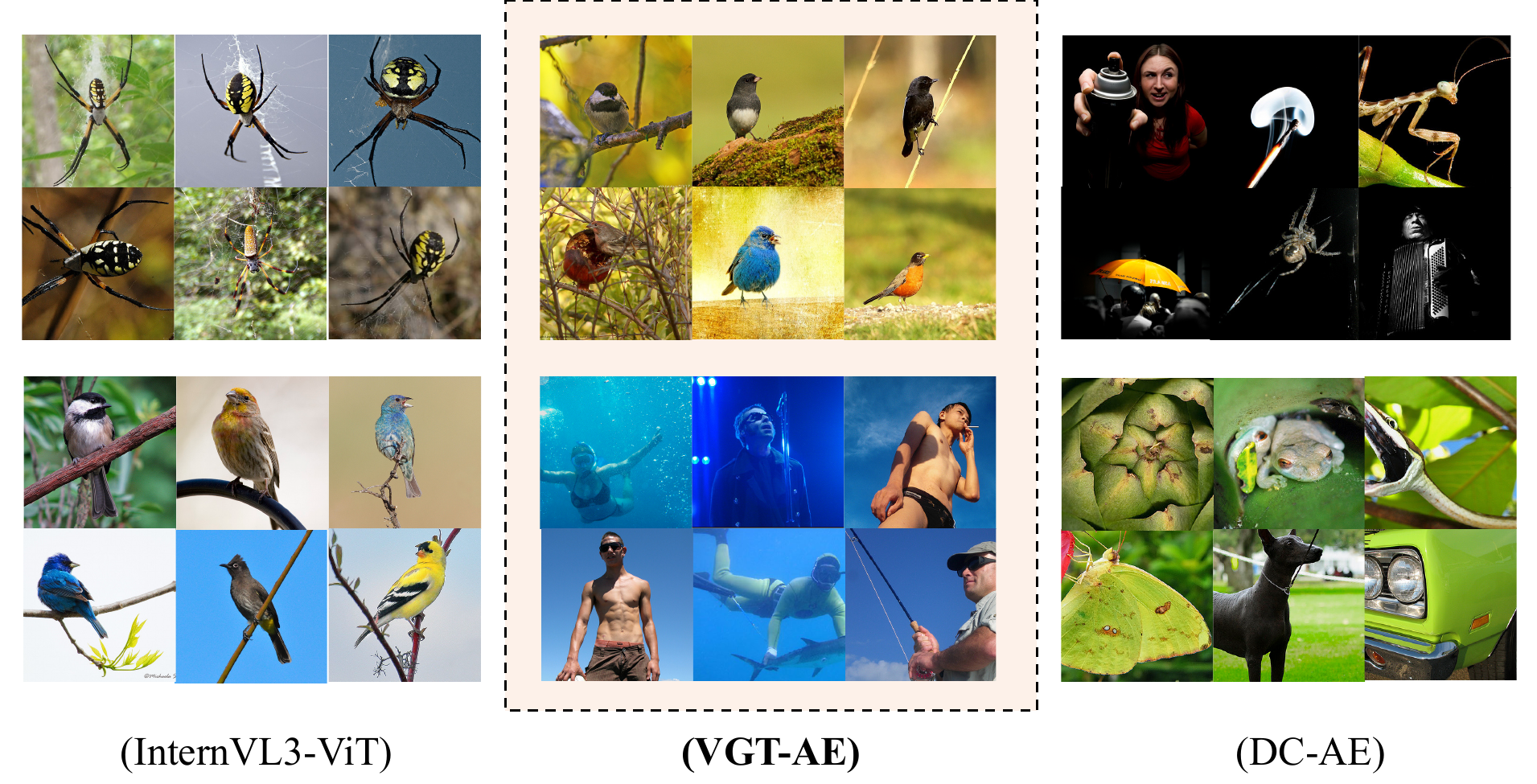}
\caption{
\textbf{Visualization of clustering results with different visual representations.} 
Our {\methodname}-AE is capable of preserving semantic structure while retaining fine-grained textures.
}
\vspace{-10pt}
\label{fig:novel_show}
\end{figure}

We further elucidate the key motivation behind the design of our visual generation tuning methods through pilot experiments. Previous studies on continuous autoregressive modeling typically adopt the pretrained VAEs derived from pixel diffusion models~\cite{mar,fluid,nextstep_1,spherear}. However, they neglect the representation misalignment with autoregressive models. Inspired by the acceleration of divergence caused by semantic alignment with the VAE latents in diffusion models~\cite{repa,rae,svg}, we perform experiments to validate the conclusion on pretrained VLMs. As presented in Tab.~\ref{tab:key_findings} Exp. 1-2, we started by fine-tuning the semantic encoders of pretraiend VLMs without feature constraints, which showcase superior capabilities of reconstruction with significant degradation of semantic structures as evidenced on the multimodal understanding benchmarks. However, a simple self-distillation loss constrains the distribution shift towards fine-grained reconstruction, while improving the understanding and generation. It indicates that the semantic structured representations facilitate the convergence of continuous AR modeling. 
Moreover, as illustrated in Figure~\ref{fig:novel_show}, we analyze the cluster results of representations of different encoders, which show that the semantic encoder (\textit{e.g., InternViT}) tends to focus on categories (high-level) while pixel encoder (\textit{e.g., DC-AE}) clusters similar textures (low-level). Our {\methodname}-AE achieves an effective compromise, producing representations that are both semantically structured and texturally rich, leading to coherent generalized clusters with refined visual detail. These observations motivate us to enhance the capabilities of visual generation by aligning the semantic encoder with the latent space of the pixel decoder for autoregressive modeling.

\subsection{{\methodname}-AE}
\label{sec:vlae}

The performance of continuous autoregressive models critically depends on the quality of visual tokenizers. Existing methods face a fundamental trade-off: traditional VAEs lack semantic supervision and suffer from variance collapse~\cite{sun2024multimodal,nextstep_1,spherear}, while directly employing structured semantic features compromises reconstruction fidelity and yields excessively high-dimensional representations~\cite{rae,unilip}, making them difficult to predict with lightweight flow matching heads. To address these limitations, we propose VGT-AE, a semantically aligned visual tokenizer trained through a two-stage paradigm that progressively optimizes for both reconstruction quality and autoregressive compatibility.

\textbf{Architecture.} VGT-AE consists of three components: (1) a semantic encoder $\mathcal{E}_{\text{vlm}}$ from a pretrained VLM (\textit{e.g.}, InternViT(vision encoder of InternVL3)~\cite{internvl3}, QwenViT(vision encoder of Qwen2.5-VL)~\cite{Qwen2_5_VL}), (2) a residual projection module $\phi$ that compresses features from dimension $d$ to $d_z = 32$, and (3) a pixel decoder $\mathcal{D}$ adapted from DC-AE~\cite{dcae}. The complete forward pass is:
\begin{equation}
\mathbf{f} = \mathcal{E}_{\text{vlm}}(\mathbf{x}), \quad \mathbf{z} = \phi(\mathbf{f}), \quad \hat{\mathbf{x}} = \mathcal{D}(\mathbf{z}).
\end{equation}

\textbf{Stage 1: Semantic-Preserving Reconstruction.} In the first stage, we jointly optimize the encoder and decoder using a composite loss that combines pixel reconstruction with semantic self-distillation:
\begin{equation}
\mathcal{L}_{\text{rec}} = \|\mathbf{x} - \hat{\mathbf{x}}\|_2^2 + \mathcal{L}_{\text{LPIPS}}(\mathbf{x}, \hat{\mathbf{x}}) + \mathcal{L}_{\text{GAN}}(\mathbf{x}, \hat{\mathbf{x}}),
\end{equation}
\begin{equation}
\mathcal{L}_{\text{distill}} = \|\mathcal{E}_{\text{teacher}}(\mathbf{x}) - \mathcal{E}_{\text{vlm}}(\mathbf{x})\|_2^2,
\end{equation}
\begin{equation}
\mathcal{L}_{\text{stage1}} = \mathcal{L}_{\text{rec}} + \lambda_{\text{distill}} \cdot \mathcal{L}_{\text{distill}},
\end{equation}
where $\|\cdot\|_2^2$ denotes mean squared error, $\mathcal{L}_{\text{LPIPS}}$~\cite{lpips} is the perceptual loss, $\mathcal{L}_{\text{GAN}}$ is the adversarial loss, $\mathcal{E}_{\text{teacher}}$ is the frozen pretrained encoder serving as the distillation target, and $\lambda_{\text{distill}}$ controls distillation strength(set to 1.0 in our experiments). This stage achieves high-fidelity reconstruction while distilling semantic structure into the compact latent space.

\textbf{Stage 2: Latent Space Regularization.} The first-stage latent distribution, while semantically structured, does not conform to the standard Gaussian prior (zero mean, unit variance) required for effective flow-based generation. Prior work~\cite{sun2024multimodal,nextstep_1,spherear} demonstrates that unregularized latent spaces pose significant challenges for flow-based methods, as mapping from simple noise distributions to complex, unconstrained target distributions becomes prohibitively difficult. Our ablation studies in Table~\ref{tab:training_ablation} confirm this limitation.

To address this, we freeze the encoder $\mathcal{E}_{\text{vlm}}$ while optimizing only the decoder $\mathcal{D}$ and projection module $\phi$, applying channel-wise layer normalization with Gaussian noise injection:
\begin{equation}
\mathbf{z}_{\text{norm}} = \frac{\mathbf{z} - \mu}{\sigma}, \quad \mathbf{z}_{\text{noisy}} = \mathbf{z}_{\text{norm}} + \epsilon,
\end{equation}
where $\mu$ and $\sigma$ represent channel-wise mean and standard deviation computed across the batch, and $\epsilon \sim \mathcal{N}(0, \sigma_{\text{noise}}^2)$ denotes injected Gaussian noise with $\sigma_{\text{noise}} = 0.1$ in our experiments. The training objective simplifies to:
\begin{equation}
\mathcal{L}_{\text{stage2}} = \|\mathbf{x} - \hat{\mathbf{x}}\|_2^2 + \mathcal{L}_{\text{LPIPS}}(\mathbf{x}, \hat{\mathbf{x}}).
\end{equation}
This regularization ensures the resulting latents maintain semantic coherence while becoming more amenable to autoregressive learning, yielding VGT-AE as a semantically meaningful and distributionally stable visual tokenizer.

\subsection{Autoregressive Modeling}
\label{sec:queryar}

Building upon the semantically structured latents from VGT-AE, we introduce QueryAR for autoregressive visual generation. While conventional raster-scan ordering is prone to error accumulation~\cite{sun2024multimodal} and existing random-order techniques~\cite{mar} enhance robustness at the cost of non-autoregressive mask modeling, QueryAR achieves flexible generation order within a standard autoregressive framework through explicit position queries.

\textbf{Causal Modeling with Position Queries.} Given a latent sequence $\mathbf{Z} = \{\mathbf{z}_1, \ldots, \mathbf{z}_N\}$ and a random permutation $\pi$, we construct the input sequence by interleaving position queries with corresponding latents: $[Q_{\pi(1)}, \mathbf{z}_{\pi(1)}, Q_{\pi(2)}, \mathbf{z}_{\pi(2)}, \ldots]$, where $Q_i \in \mathbb{R}^{d_{\text{model}}}$ are learnable position embeddings. The language model learns the conditional distribution:
\begin{equation}
p_{\theta}(\mathbf{z}_{\pi(t)} \mid \mathbf{H}_{<t}, Q_{\pi(t)}),
\end{equation}
where $\mathbf{H}_{<t}$ denotes the causal context containing all previously generated latents and their associated position queries. This formulation maintains autoregressive properties while supporting flexible generation orders.

\textbf{Continuous Latent Modeling via Flow Matching.} For modeling continuous latents, we employ a lightweight flow matching head~\cite{mar} conditioned on language model hidden states. The language model processes input sequences to produce hidden states $\mathbf{H} = f_{\theta}([Q_{\pi(1)}, \mathbf{z}_{\pi(1)}, \ldots])$, which serve as conditioning signals for the flow matching procedure. The training objective follows the standard formulation:
\begin{equation}
\mathcal{L}_{\text{fm}} = \mathbb{E}_{t,\epsilon} \left[\| (\mathbf{z}_{\text{target}} - \epsilon) - v_{\theta}(\mathbf{z}_t, t, \mathbf{H}) \|_2^2\right],
\end{equation}
where $\mathbf{z}_t = t \cdot \mathbf{z}_{\text{target}} + (1-t) \cdot \epsilon$ is the linearly interpolated latent at timestep $t \in [0,1]$, $\epsilon \sim \mathcal{N}(0, I)$ is Gaussian noise, and $v_{\theta}$ is the predicted vector field. For timestep scheduling, we implement the dimensional-adaptive strategy from~\cite{noisetimestep} with $t_m = \frac{\alpha t_n}{1 + (\alpha - 1)t_n}$, where $\alpha = \sqrt{m/n}$ and $n=4096$ serves as the reference dimension.

\textbf{Parallel Decoding for Efficient Inference.} The position-query mechanism substantially enhances inference efficiency through partially parallel decoding. Given previously generated latents $\mathbf{z}_{1:k}$ and target position queries $Q_{k+1:k+m}$, the language model processes these inputs to produce hidden states for the subsequent $m$ positions in a single forward pass:
\begin{equation}
\mathbf{H}_{k+1:k+m} = f_{\theta}(\mathbf{z}_{1:k}, Q_{k+1:k+m}).
\end{equation}
These hidden states then condition the flow matching head for deterministic sampling:
\begin{equation}
\hat{\mathbf{Z}}_{k+1:k+m} = \text{FlowSample}(\mathbf{H}_{k+1:k+m}, \epsilon_{\theta}),
\end{equation}
where $\epsilon_{\theta}$ denotes the flow matching head. Our ablation study in Section~\ref{sec:ablation_queryar} validates this design, demonstrating that multiple tokens can be generated simultaneously while preserving generation quality through the position-query mechanism.

\section{Experiments}
\label{sec:experiments}

\subsection{Experimental Setups}
\label{sec:setting}

\textbf{Datasets.}
{\methodname} is pretrained on the BLIP3-o~\cite{blip3-o} open-source dataset, which comprises 27M samples recaptioned by Qwen2.5-VL-7B~\cite{Qwen2_5_VL}, 5M samples from CC12M~\cite{cc12m}, and 4M synthesized images from JourneyDB~\cite{journeydb}. For supervised fine-tuning, we use a total of 200K high-quality aesthetic samples collected from ShareGPT-4o~\cite{sharegpt_4o}, Echo-4o~\cite{echo4o}, and BLIP3-o~\cite{blip3-o}, to enhance visual quality and prompt alignment.

\begin{table}[h!]
\centering
\scalebox{0.8}{
    \begin{tabular}{c|c|ccc}
    \toprule
    \rowcolor{white}\textbf{Method} & \textbf{Ratio} & \textbf{rFID$\downarrow$} & \textbf{PSNR$\uparrow$} & \textbf{SSIM$\uparrow$} \\
    \midrule
    \multicolumn{5}{c}{\textit{Generative Only Tokenizer}} \\
    \midrule
    VQGAN~\cite{vqgan} & 16 & 4.98 & 20.00 & 0.629 \\
    LlamaGen~\cite{llamagen} & 16 & 2.19 & 20.79 & 0.675 \\
    VAR~\cite{var} & 16 & 1.00 & 22.63 & 0.755 \\
    Open-MAGVIT2~\cite{open-magvit2} & 16 & 1.67 & 22.70 & 0.640 \\
    RAE~\cite{rae} & 16 & 0.49 & 19.23 & 0.620 \\
    SD-VAE~\cite{kl-16} & 16 & 2.64 & 22.13 & 0.590 \\
    DC-AE~\cite{dcae} & 32 & 0.69 & 23.85 & 0.660 \\
    \midrule
    \multicolumn{5}{c}{\textit{CLIP-based Tokenizer}} \\
    \midrule
    VILA-U~\cite{Vila-u} & 16 & 1.80 & - & - \\
    TokenFlow~\cite{tokenflow} & 16 & 1.37 & 21.41 & 0.687 \\
    DualViTok~\cite{Dualtoken} & 16 & 1.37 & 22.53 & 0.741 \\
    UniLIP~\cite{unilip} & 32 & 0.79 & 22.99 & 0.747 \\
    \midrule
    \multicolumn{5}{c}{\textit{Ours}} \\
    \midrule
    \rowcolor{cyan!10} \textbf{{\methodname}-AE (QwenViT)} & 28 & 1.93 & 20.12 & 0.677 \\
    \rowcolor{cyan!10} \textbf{{\methodname}-AE (InternViT)} & 28 & \textbf{0.50} & \textbf{26.67} & \textbf{0.863} \\
    \bottomrule
    \end{tabular}
}
\caption{
Comparisons of reconstruction quality on the 256 $\times$ 256 ImageNet 50k validation set. 
Ratio indicates the downsample ratio relative to the original image.
Best results are highlighted in bold.{\methodname}-AE (QwenViT) and {\methodname}-AE (InternViT) use the vision encoders of Qwen2.5-VL and InternVL3, respectively.
}
\label{tab:tokenizer}
\vspace{-15pt}
\end{table}

\hspace{-0.4cm}\textbf{Implementation Details.}  
We develop {\methodname} models with 0.6B and 1.6B parameters, instantiated from the Qwen2.5-VL~\cite{Qwen2_5_VL} and InternVL3~\cite{internvl3} VLM families. {\methodname}-AE adopts the vision encoders of Qwen2.5-VL (QwenViT) and InternVL3 (InternViT) and pairs them with the DC-AE decoder~\cite{dcae}. For autoregressive visual generation, QueryAR uses {\methodname}-AE as the visual tokenizer and shares the language model of the underlying VLM.
 To achieve target model sizes, we apply pruning to the Qwen2.5-VL models. We pretrain using the AdamW optimizer with a batch size of 256 and a learning rate of \(2.0 \times 10^{-4}\), with cosine learning-rate decay. The optimizer settings are \(\beta = (0.9, 0.95)\) and weight decay 0.05. EMA is applied with a decay rate of 0.9999. Models are pretrained for 100K steps, and ablation experiments are conducted for 50K steps. Fine-tuning for autoregressive visual generation is performed for 5,000 iterations with a learning rate of \(5 \times 10^{-5}\). We use a causal mask with a maximum sequence length of 1024 tokens.

\hspace{-0.4cm}\textbf{Evaluation Metrics.}
We assess reconstruction quality using rFID, PSNR, and SSIM on the ImageNet-1K validation set. For visual generation, we evaluate on GenEval~\cite{geneval} and DPG-Bench~\cite{dpgbench}. For multimodal understanding, we evaluate on MME-P: MME-Perception~\cite{mme}, MMB: MMBench-en~\cite{mmbench}, TQA: TextVQA~\cite{textvqa} and AI2D~\cite{ai2d}.

\subsection{Visual Tokenizer}
\label{sec:exp_tokenzier}

\begin{figure}
    \centering
    \includegraphics[width=1\linewidth]{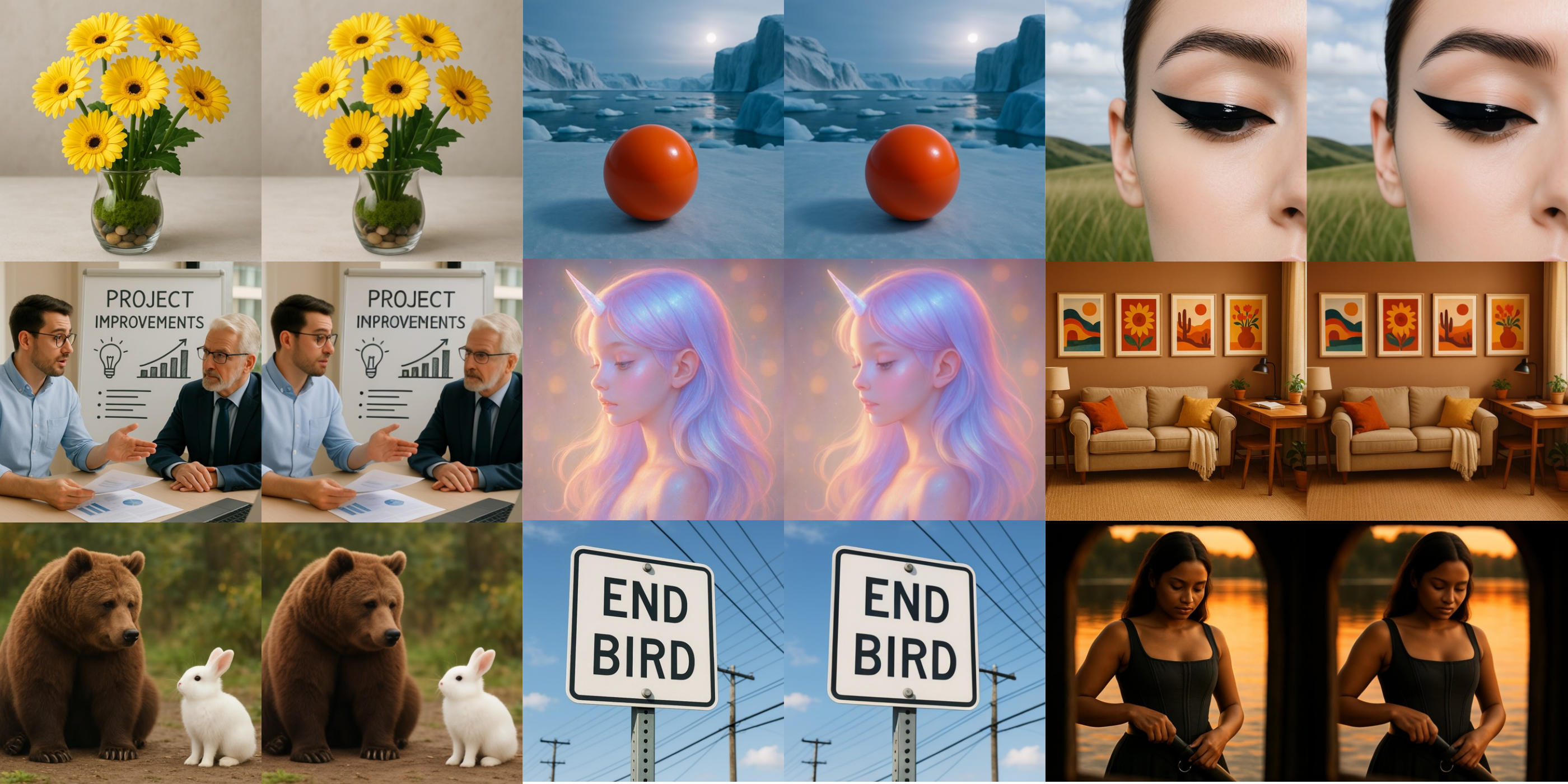}
    \caption{
    Visualization of reconstruction results from our {\methodname}-AE-InternViT. \textbf{Left}: input image; \textbf{Right}: reconstructed image.
    }
    \label{fig:rec_show}
    \vspace{-10pt}
\end{figure}

{\methodname}-AE, a semantically aligned visual tokenizer, successfully bridges the powerful representational capabilities of vision encoders from VLMs with the demands of high-fidelity image reconstruction through a carefully designed two-stage training strategy. Experimental results in Tab.~\ref{tab:tokenizer} demonstrate that our two-stage approach enables high-quality reconstruction, achieving state-of-the-art reconstruction performance. Notably, {\methodname}-AE-InternViT attains remarkable metrics on the ImageNet 256$\times$256 validation set, with rFID of \textbf{0.50}, PSNR of \textbf{26.67}, and SSIM of \textbf{0.863}, significantly surpassing existing generative-only and CLIP-based tokenizers. We provide qualitative visualizations in Fig.~\ref{fig:rec_show}.

\begin{table*}[t]
\centering
\scalebox{0.7}{
\begin{tabular}{crr|ccccc|rrrrr}
\toprule
\multirow{2}{*}{\textbf{Method}} & \multirow{2}{*}{\textbf{Data.}} & \multirow{2}{*}{\textbf{\# Params.}} & \multicolumn{5}{c}{\textbf{GenEval}~\cite{geneval}} & \multicolumn{5}{c}{\textbf{DPG-Bench}~\cite{dpgbench}} \\
\cmidrule(lr){4-8} \cmidrule(lr){9-13}
& & & \textbf{Two Object} & \textbf{Counting} & \textbf{Colors} & \textbf{Position} & \textbf{Overall$\uparrow$} & \textbf{Global} & \textbf{Attribute} & \textbf{Relation} & \textbf{Other} & \textbf{Overall$\uparrow$} \\
\midrule
\multicolumn{13}{c}{\textit{Diffusion-based Model}} \\
\midrule
SDv1.5~\cite{sd} & \textgreater2000M & 0.9B & 0.38 & 0.35 & 0.76 & 0.04 & 0.43 & 74.63 & 75.39 & 73.49 & 67.81 & 63.18 \\
PixArt-$\alpha$~\cite{pixartalpha} & 25M & 0.6B & 0.50 & 0.44 & 0.80 & 0.08 & 0.48 & 74.97 & 78.60 & 82.57 & 76.96 & 71.11 \\
SDv2.1~\cite{sd} & - & 0.9B & 0.51 & 0.44 & 0.85 & 0.07 & 0.50 & - & - & - & - & - \\
SDXL~\cite{sdxl} & - & 2.6B & 0.74 & 0.39 & 0.85 & 0.15 & 0.55 & 83.27 & 80.91 & 86.76 & 80.41 & 74.65 \\
DALLE3~\cite{dalle2} & - & - & 0.87 & 0.47 & 0.83 & \textbf{0.43} & 0.67 & \textbf{90.97} & 88.39 & \textbf{90.58} & \textbf{89.83} & 83.50 \\
SD3-Medium~\cite{sd} & - & 2B & \textbf{0.94} & \textbf{0.72} & \textbf{0.89} & 0.33 & \textbf{0.74} & 87.90 & \textbf{88.83} & 80.70 & 88.68 & \textbf{84.08} \\
\midrule
\multicolumn{13}{c}{\textit{Autoregressive-based Model}} \\
\midrule
Chameleon~\cite{chameleo} & \textgreater1.4B & 7B & - & - & - & - & 0.39 & - & - & - & - & - \\
Fuild~\cite{fluid} & 2048M & 0.4B& - & - & - & - & 0.45 & - & - & - & - & - \\
Fuild~\cite{fluid} & 2048M & 0.7B & - & - & - & - & 0.51 & - & - & - & - & - \\
LlamaGen~\cite{llamagen} & - & 0.8B & 0.34 & 0.21 & 0.58 & 0.07 & 0.32 & 81.76 & 76.17 & 84.76 & 58.40 & 64.84 \\
EMU3-Gen~\cite{emu3} & - & 8B & 0.71 & 0.34 & 0.81 & 0.17 & 0.54 & 85.21 & 86.84 & \textbf{90.22} & 83.15 & 80.60 \\
TokenFlow~\cite{tokenflow} & - & 13B & 0.66 & 0.40 & 0.84 & 0.17 & 0.55 & 78.72 & 81.29 & 85.22 & 71.20 & 73.38 \\
SEED-X~\cite{seedx} & - & 0.7B & 0.65 & 0.31 & 0.80 & 0.18 & 0.51 & - & - & - & - & - \\
Janus~\cite{janus} & 198M & 1.3B & 0.68 & 0.30 & 0.84 & 0.46 & 0.61 & 82.33 & 87.70 & 85.46 & 86.41 & 79.68 \\
SimpleAR~\cite{simplear} & - & 1.5B & \textbf{0.90} & - & - & 0.28 & 0.63 & \textbf{87.97} & - & 88.33 & - & 81.97 \\
VAR~\cite{var} & - & - & - & - & - & - & 0.53 & - & - & - & - & 71.08 \\
Janus-Pro~\cite{janus-pro} & 198M & 1B & 0.82 & 0.51 & \textbf{0.89} & 0.65 & 0.73 & 87.58 & \textbf{88.17} & 88.98 & \textbf{88.30} & 82.63 \\
NextStep-1~\cite{nextstep_1} & \textgreater2048M & 14B & - & - & - & - & 0.63 & - & - & - & - & \textbf{85.28} \\
\rowcolor{cyan!10} \textbf{{\methodname} (InternVL3)} & \textbf{$<$25M} & 0.6B & 0.76 & 0.51 & 0.78 & 0.70 & 0.71 & 81.55 & 82.96 & 85.91 & 82.21 & 72.01 \\
\rowcolor{cyan!10} \textbf{{\methodname} (InternVL3)} & \textbf{$<$25M} & 1.6B & 0.83 & \textbf{0.61} & 0.84 & 0.70 & 0.75 & 86.55 & 83.76 & 84.36 & 85.14 & 74.43 \\
\rowcolor{cyan!10} \textbf{{\methodname} (Qwen2.5-VL)} & \textbf{$<$25M} & 0.6B & 0.78 & 0.54 & 0.83 & 0.68 & 0.72 & 85.16 & 84.89 & 85.50 & 82.00 & 75.18 \\
\rowcolor{cyan!10} \textbf{{\methodname} (Qwen2.5-VL)} & \textbf{$<$25M} & 1.6B & 0.85 & 0.59 & 0.87 & \textbf{0.74} & \textbf{0.77} & 87.41 & 85.67 & 86.78 & 87.73 & 78.73 \\
\bottomrule
\end{tabular}
}
\caption{
Comparisons of visual generation quality on GenEval~\cite{geneval} and DPG-Bench~\cite{dpgbench}. Bold numbers indicate the best performance in each column. Our {\methodname} models demonstrate highly competitive results, with the VGT-Qwen2.5-VL variant establishing new state-of-the-art performance on GenEval while trained on remarkably limited data (only 25M samples).
}
\label{tab:generation_comparison}
\vspace{-8pt}
\end{table*}

\subsection{Visual Generation}
\label{sec:exp_generation}

\begin{figure}[ht!]
    \centering
    \includegraphics[width=1\linewidth]{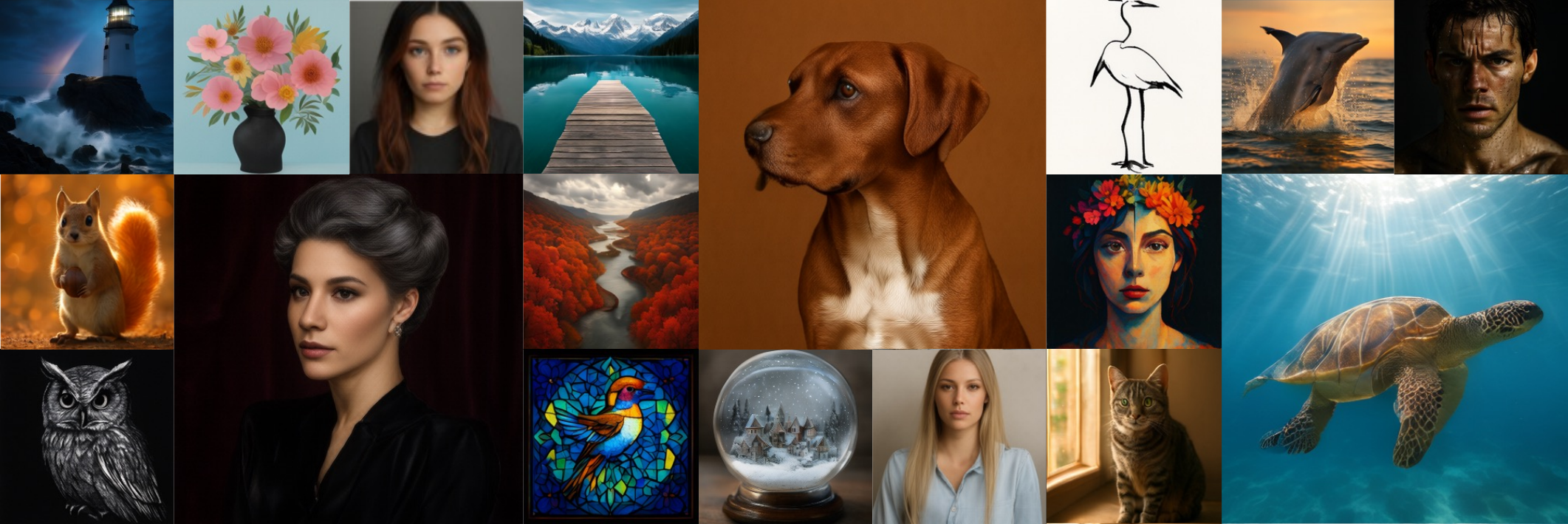}
    \caption{Our \textbf{{\methodname}-1.6B} based on Qwen2.5-VL and InternVL3, endow pretrained vision--language models trained on multimodal understanding tasks with high-quality visual generation, enabling them to produce diverse and realistic images.}
    \label{fig:show_case}
    \vspace{-10pt}
\end{figure}

We systematically evaluate the visual generation performance of our proposed {\methodname} paradigm on two established benchmarks: GenEval~\cite{geneval} and DPG-Bench~\cite{dpgbench}. As shown in Table~\ref{tab:generation_comparison}, our method demonstrates strong performance, particularly on the GenEval benchmark. As demonstrated in Figure~\ref{fig:show_case}, our {\methodname} paradigm effectively endows well-pretrained vision--language models with the ability to produce diverse and realistic high-quality images. The {\methodname} model based on Qwen2.5-VL~\cite{Qwen2_5_VL} achieves a state-of-the-art overall score of \textbf{0.77}, surpassing existing autoregressive generation models including Janus-Pro~\cite{janus-pro}, TokenFlow~\cite{tokenflow}, and SimpleAR~\cite{simplear}. Notably, it also competes favorably with or even exceeds several large diffusion models such as SDXL~\cite{sdxl}, SD3-Medium~\cite{sd}, and DALLE3~\cite{dalle2} across multiple sub-metrics. This strong performance is achieved with only 25M training samples, highlighting {\methodname}'s superior instruction-following ability and its effectiveness in constrained generation. A key factor behind this improvement is our visual representation module, {\methodname}-AE, which builds on representations learned by large vision--language models. This design offers inherent vision--language alignment, allowing the generator to maintain strong semantic consistency with textual instructions while avoiding the structural information loss commonly seen in traditional VAEs.

Our results challenge long-standing assumptions in visual generation. Conventional autoregressive models typically require orders of magnitude more training data (hundreds of millions to billions of samples) yet still exhibit unstable performance. In contrast, {\methodname} shows clear advantages on challenging constraint-based tasks such as \emph{Counting} and \emph{Position}. More importantly, {\methodname} achieves these improvements while maintaining substantially higher training efficiency compared to diffusion-based approaches. For instance, SD3-Medium~\cite{sd} relies on 2B parameters and expensive training procedures to achieve high GenEval scores, whereas our {\methodname}-Qwen2.5-VL-1.6B model attains comparable performance using only 25M training samples. This challenges the long-held belief that autoregressive models inherently produce lower-quality images than diffusion models of comparable scale. Even our smaller {\methodname}-0.6B model surpasses larger systems such as NextStep-1~\cite{nextstep_1} and Janus-Pro~\cite{janus-pro} on several metrics, further confirming that the vision--language aligned representation in {\methodname} enables stronger generalization and more controllable visual generation. These results point to a promising new paradigm for efficient yet high-quality autoregressive visual generation.

\subsection{Ablation Studies}
\label{sec:ablation}
To comprehensively evaluate the design choices in {\methodname}, we conduct ablation studies across five critical components: the {\methodname}-AE decoder architecture (Section~\ref{sec:ablation_vlvae}), training methodology (Section~\ref{sec:ablation_reg}), the reconstruction-generation trade-off (Section~\ref{sec:ablation_recVsGen}), cross-architecture compatibility (Section~\ref{sec:ablation_different_llms}), and QueryAR structure (Section~\ref{sec:ablation_queryar}).

\subsubsection{{\methodname}-AE Architecture}
\label{sec:ablation_vlvae}
\begin{table}[h!]
\centering
\scalebox{0.75}{
    \begin{tabular}{c|ccc|c}
    \toprule
    \textbf{Decoder Type} & \textbf{rFID$\downarrow$} & \textbf{PSNR$\uparrow$} & \textbf{SSIM$\uparrow$} & \textbf{Param.} \\
    \midrule
    \multicolumn{5}{c}{\textit{{\methodname}-AE with Qwen2.5-VL Vision Encoder}} \\
    \midrule
    ViT & 17.32 & 18.56 & 0.518 & 845M \\
    SD-VAE & 2.81 & 19.22 & 0.640 & \textbf{743M} \\
    DC-AE$^\dagger$ & 2.71 & 19.41 & \textbf{0.647} & 853M \\
    \rowcolor{blue!10} \textbf{DC-AE} & \textbf{2.67} & \textbf{19.68} & 0.563 & 853M \\
    \midrule
    \multicolumn{5}{c}{\textit{{\methodname}-AE with InternVL3 Vision Encoder}} \\
    \midrule
    ViT & 15.80 & 20.01 & 0.611 & 464M \\
    SD-VAE & 1.32 & 23.10 & \textbf{0.816} & \textbf{362M} \\
    DC-AE$^\dagger$ & 1.14 & 23.86 & 0.792 & 472M \\
    \rowcolor{blue!10} \textbf{DC-AE} & \textbf{1.13} & \textbf{23.95} & 0.801 & 472M \\
    \bottomrule
    \end{tabular}
}
\caption{
Ablation study on decoder architectures for {\methodname}-AE with different vision-language model encoders. $^\dagger$ indicates models trained without pretrained weights.
}
\label{tab:decoder_ablation}
\vspace{-8pt}
\end{table}

In Table~\ref{tab:decoder_ablation}, we compare several mainstream decoder architectures for {\methodname}-AE: the high-compression DC-AE from SANA~\cite{dcae}, the SD-VAE decoder from Stable Diffusion~\cite{kl-16}, and a ViT-based decoder of comparable parameter scale. Results show that both DC-AE and SD-VAE decoders demonstrate significant advantages regardless of whether Qwen2.5-VL or InternVL3 vision encoders are employed, while the ViT decoder substantially lags behind across all metrics.

For instance, in the InternVL3 configuration, DC-AE reduces rFID from 15.80 (ViT) to \textbf{1.13} and improves PSNR to \textbf{23.95}. The relatively small performance gap between DC-AE and SD-VAE (1.13 vs. 1.32 rFID) indicates that both effectively recover image details while maintaining semantic consistency. Based on these findings, we select DC-AE as our default decoder architecture to achieve the optimal balance between semantic structure preservation, visual detail restoration, and parameter efficiency.

\subsubsection{{\methodname}-AE Training Methodology}
\label{sec:ablation_reg}

\begin{table}[h!]
\centering
\scalebox{0.85}{
\begin{tabular}{c|ccc|ccc}
\toprule
\multirow{2}{*}{\textbf{Exp.}} & \multicolumn{3}{c|}{\textbf{Training Strategy}} & \multirow{2}{*}{\textbf{rFID$\downarrow$}} & \multirow{2}{*}{\textbf{PSNR$\uparrow$}} & \multirow{2}{*}{\textbf{Geneval$\uparrow$}} \\
\cmidrule(lr){2-4}
 & \textbf{Stage} & \textbf{Norm} & \textbf{Noise} & & & \\
\midrule
1 & Stage1 & \ding{55} & \ding{55} & \textbf{0.98} & \textbf{24.30} & 0.36 \\
2 & Stage2 & \ding{51} & \ding{55} & 1.05 & 24.21 & 0.52 \\
\rowcolor{blue!10} 3 & Stage2 & \ding{51} & 0.1 & 1.13 & 23.95 & \textbf{0.54} \\
\bottomrule
\end{tabular}
}
\caption{
Ablation study on {\methodname}-AE's training strategies. Stage1 focuses on reconstruction, while Stage2 enhances generation through latent space regularization. The performance progression from Experiment 1 to 3 underscores the importance of regularization for learning generation-friendly representations.
}
\label{tab:training_ablation}
\end{table}

Our ablation study provides key findings regarding {\methodname}-AE's training paradigm. As summarized in Table~\ref{tab:training_ablation}, Stage 1 training achieves the best reconstruction quality but yields poor generation performance (GenEval=0.36). This suggests that latents optimized solely for reconstruction are suboptimal for the denoising process in autoregressive generation. Introducing normalization in Stage 2 (Experiment 2) improves generation substantially, and further adding mild noise injection (Experiment 3) yields the best generation result (GenEval=\textbf{0.54}), despite a slight degradation in reconstruction metrics.

\begin{figure}[h!]
\centering
\includegraphics[width=1.0\linewidth]{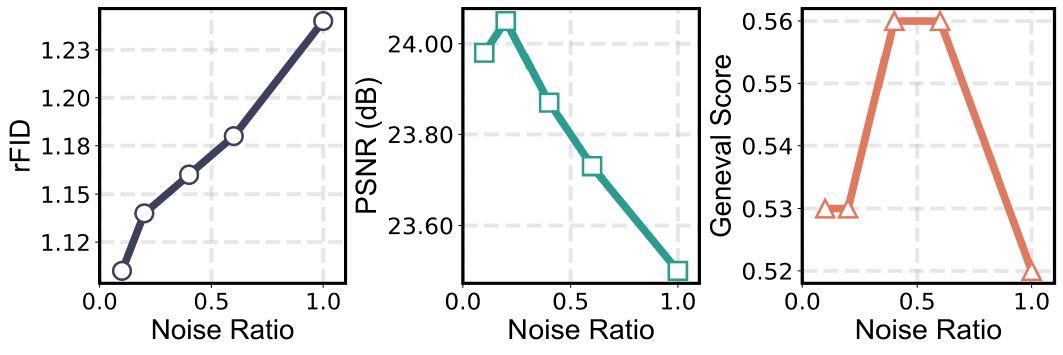}
\caption{
Impact of noise regularization on the reconstruction-generation trade-off. While reconstruction quality (rFID, PSNR) degrades with increasing noise, generation performance (GenEval) peaks at moderate noise intensities ($\sigma=0.4$--$0.6$).
}
\label{fig:noise_ablation}
\vspace{-5.0pt}
\end{figure}

Figure~\ref{fig:noise_ablation} further elucidates this trade-off. Reconstruction quality decreases monotonically with noise level, whereas generation performance peaks at moderate levels ($\sigma=0.4$-$0.6$). This demonstrates that an appropriate level of regularization is crucial for striking a balance.

Collectively, these results validate {\methodname}-AE's two-stage training strategy as an effective paradigm for learning latent representations that reconcile high-fidelity reconstruction with effective autoregressive generation.

\subsubsection{Reconstruction vs. Generation}
\label{sec:ablation_recVsGen}

\begin{table}[!t]
\centering

\scalebox{0.7}{
\begin{tabular}{lcccc}
\toprule
\textbf{Model} & \textbf{rFID$\downarrow$} & \textbf{PSNR$\uparrow$} & \textbf{Geneval$\uparrow$} & \textbf{DPG-Bench$\uparrow$} \\
\midrule
VGT(Qwen2.5-VL) & 1.93 & 20.12 & \textbf{0.72} & \textbf{75.18} \\
VGT(InternVL3) & \textbf{0.50} & \textbf{26.67} & 0.71 & 72.01 \\
VGT(InternVL3, Low Rec.) & 1.13 & 23.95 & \textbf{0.72} & 73.50 \\
\bottomrule
\end{tabular}
}
\caption{Comparison of reconstruction and generation performance across VGT models(0.6B). Reconstruction-oriented metrics (rFID, PSNR) and generation-oriented benchmarks (Geneval, DPG-Bench) exhibit a consistent inverse trend, highlighting a structural trade-off inherent to autoregressive latent modeling.}
\label{tab:rec_gen_tradeoff}
\end{table}

\begin{figure}[!ht]
\centering
\includegraphics[width=1.0\linewidth]{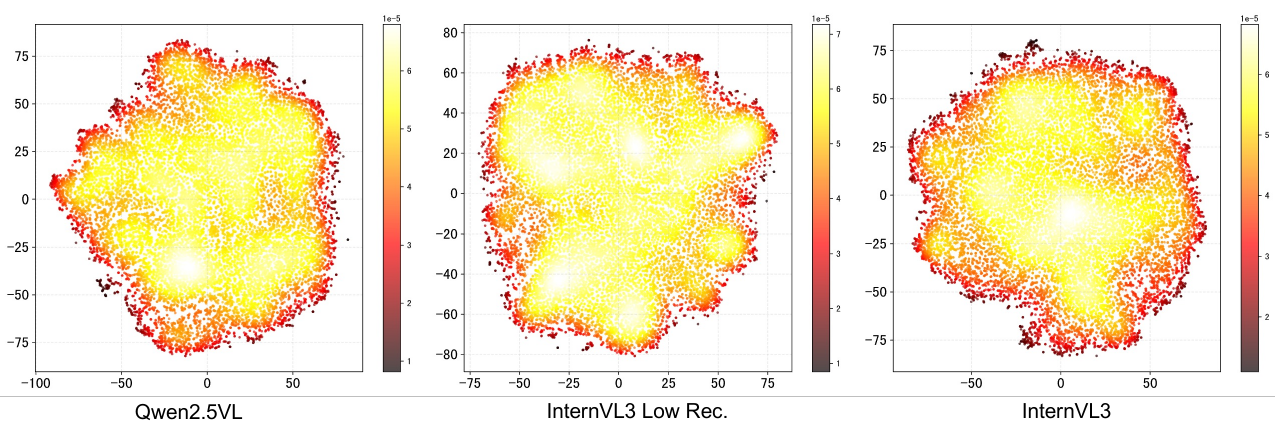}
\caption{t-SNE of latent tokens from three {\methodname}-AE variants. Generation-oriented models (Qwen2.5-VL and InternVL3 Low Rec.) yield dispersed, semantically structured manifolds, while the reconstruction-oriented InternVL3 forms compact, entangled clusters, reflecting different requirements on the latent space for reconstruction vs.\ autoregressive generation.}
\label{fig:tsne_latents}
\end{figure}

\begin{figure}[!ht]
\centering
\includegraphics[width=1.0\linewidth]{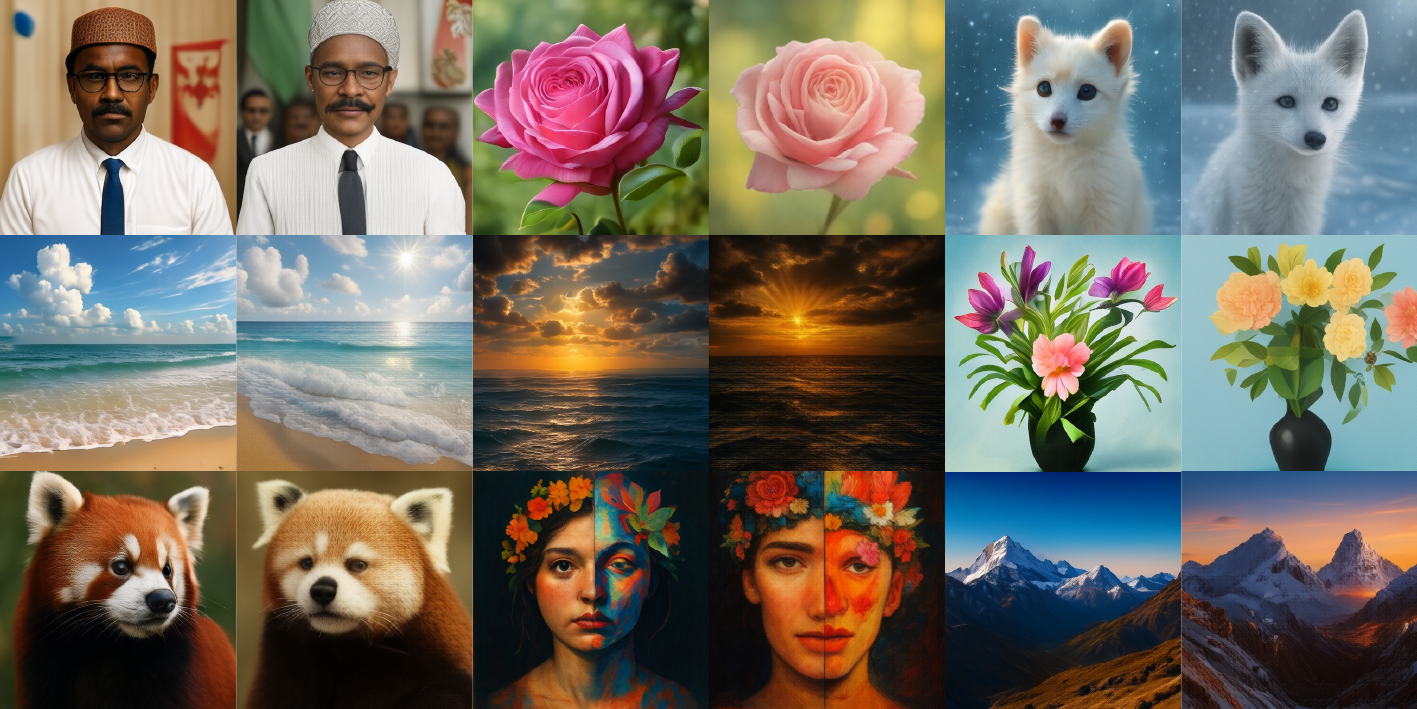}
\caption{
Qualitative comparison between InternVL3-based {\methodname} (\textbf{left}) and Qwen2.5-VL-based {\methodname} (\textbf{right}). 
The InternVL3 variant, despite lower generation metrics, produces noticeably sharper illumination and more detailed textures.
}
\label{fig:internvlvsqwen}
\end{figure}

We study the trade-off between reconstruction fidelity and generative capability using three {\methodname}-AE variants instantiated with Qwen2.5-VL~\cite{Qwen2_5_VL} and InternVL3~\cite{internvl3}. As shown in Table~\ref{tab:rec_gen_tradeoff}, the reconstruction-oriented InternVL3 variant, which yields the most compact and pixel-aligned latent representations, achieves the best reconstruction scores but underperforms on GenEval~\cite{geneval} and DPG-Bench~\cite{dpgbench}. Relaxing its reconstruction objective (InternVL3 Low Rec.) degrades pixel-level fidelity yet consistently improves generation metrics, and the Qwen2.5-VL-based {\methodname} shows a similar trend favoring generation. Figure~\ref{fig:tsne_latents} further indicates that these gains correlate with more dispersed, semantically separated manifolds, whereas the reconstruction-oriented variant forms dense, entangled clusters. Qualitatively (Figure~\ref{fig:internvlvsqwen}), however, the high-reconstruction InternVL3-based {\methodname} still produces sharper illumination and finer textures than its Qwen2.5-VL counterpart, indicating that higher automatic generation metrics (e.g., GenEval and DPG-Bench) are not necessarily aligned with perceptual sharpness.

This trade-off echoes findings in diffusion transformers, where semantically structured representations (e.g., CLIP-/DINO-like features) improve stability and sample quality at the expense of exact pixel matching~\cite{rae,svg,unilip}, and aligns with observations in discrete visual tokenization~\cite{tokenflow, Dualtoken, vqkd}. Together, these results highlight that \emph{representation design} is central to visual generation: latents optimized purely for pixel reconstruction are overly constrained for autoregressive prediction, whereas semantically organized, weakly coupled latents are more favorable for generation~\cite{nextstep_1, vavae}. {\methodname}-AE targets this regime via a two-stage training scheme: self-distillation to preserve the VLM encoder’s semantic structure, followed by channel normalization and noise regularization~\cite{sun2024multimodal,nextstep_1} to enhance robustness and reconstruction quality. This allows {\methodname} to approach the reconstruction fidelity of specialized VAEs while retaining a latent space intrinsically well-suited for continuous-space autoregressive generation.

\subsubsection{Different Language Models in {\methodname}}
\label{sec:ablation_different_llms}

We study the coupling between the autoencoder (AE) and the language model (LLM) by evaluating four {\methodname} configurations: (i) AE and LLM both from Qwen2.5-VL, (ii) both from InternVL3, (iii) a Qwen2.5-VL AE with an InternVL3 LLM, and (iv) a VAE-based DC-AE baseline. As summarized in Table~\ref{tab:different_llms_vgt}, using AE and LLM from the same VLM family yields the strongest overall performance, reflecting coherent cross-modal alignment from large-scale multimodal pre-training. The mismatched configuration (Qwen2.5-VL AE + InternVL3 LLM) still surpasses the VAE baseline on generation metrics, while the VAE lags behind across all evaluations, confirming that pixel-reconstruction-oriented latents are poorly suited for autoregressive decoding~\cite{kl-16,dcae}.

\begin{table}[!ht]
\centering
\scalebox{0.7}{
\begin{tabular}{lccc}
\toprule
\textbf{Autoencoder} & \textbf{LLM From} & \textbf{Geneval$\uparrow$} & \textbf{DPG-Bench$\uparrow$} \\
\midrule
VGT-AE (Qwen2.5-VL) & Qwen2.5-VL & \textbf{0.77} & \textbf{78.73} \\
VGT-AE (InternVL3) & InternVL3 & 0.75 & 74.43 \\
VGT-AE (Qwen2.5-VL) & InternVL3 (mismatch) & 0.75 & 72.46 \\
DC-AE (VAE baseline) & InternVL3 & 0.64 & 67.20 \\
\bottomrule
\end{tabular}
}
\caption{Performance comparison across different Autoencoder and LLM combinations within the VGT paradigm. 
Even in mismatched settings, VGT-AE significantly outperforms the VAE baseline (DC-AE), demonstrating the robustness and generative friendliness of VGT latent representations.}
\label{tab:different_llms_vgt}
\end{table}

These observations support a central insight of the {\methodname} paradigm: visual representations learned from multimodal understanding data can be transferred to language models for autoregressive visual generation with a \emph{low alignment cost}. Because {\methodname}-AE is trained inside a multimodally aligned VLM (Section~\ref{sec:exp_tokenzier}), its latent tokens inherit semantic structure that is naturally compatible with different LLMs, with matched AE–LLM pairs amplifying this effect and mismatched pairs still outperforming VAEs that must be aligned from scratch. Thus, {\methodname} effectively upgrades existing VLMs into unified models for both multimodal understanding and visual generation, and even under imperfect component pairing (Table~\ref{tab:different_llms_vgt}), its semantically organized latent space makes bridging to autoregressive generation substantially easier than with conventional VAE-based approaches.

\subsubsection{QueryAR vs. MAR}
\label{sec:ablation_queryar}
\begin{table}[h!]
\centering
\scalebox{0.80}{
\begin{tabular}{c|cc|cc}
\toprule
\textbf{\# Exp.} & \textbf{Method} & \textbf{Accel. Ratio} & \textbf{Geneval$\uparrow$} & \textbf{DPGBench$\uparrow$} \\
\midrule
1 & MAR & $1\times$ & \textbf{0.51} & 74.39 \\
2 & MAR & $4\times$ & 0.46 & \textbf{75.29} \\
\rowcolor{blue!10} 3 & QueryAR & $1\times$ & \textbf{0.62} & 77.78 \\
\rowcolor{blue!10} 4 & QueryAR & $4\times$ & 0.59 & \textbf{78.14} \\
\bottomrule
\end{tabular}
}
\caption{
Comparative analysis of decoding strategies: QueryAR vs. MAR across different acceleration ratios. QueryAR maintains competitive performance even at 4$\times$ acceleration, demonstrating its robustness for parallel decoding.
}
\label{tab:queryar_ablation}
\end{table}

To rigorously evaluate the efficacy of QueryAR, we conduct a comparative analysis against MAR~\cite{mar}, a prominent random-order autoregressive method, using the {\methodname}-AE Qwen2.5-VL model. As shown in Table~\ref{tab:queryar_ablation}, the performance of MAR degrades under acceleration, with the GenEval score dropping from 0.51 ($1\times$) to 0.46 ($4\times$). This indicates a compromise in overall generation quality under increased acceleration.

In stark contrast, QueryAR consistently outperforms MAR across all tested acceleration ratios. At $1\times$ decoding, QueryAR achieves a superior 0.62 GenEval score. Crucially, at $4\times$ acceleration, QueryAR maintains a strong GenEval of 0.59 and achieves the highest DPG-Bench score (78.14) in this comparison. This robust performance demonstrates QueryAR's ability to achieve significant inference speed-ups without compromising generation quality.

\begin{figure}
    \centering
    \includegraphics[width=1.0\linewidth]{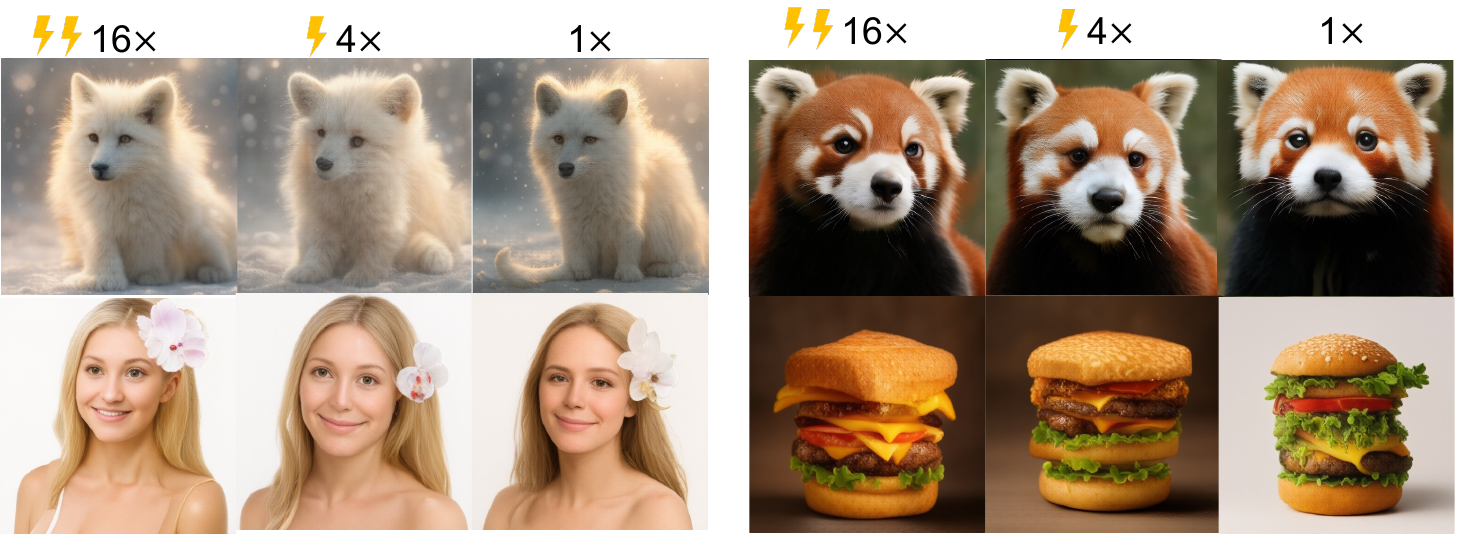}
    \caption{\textbf{QueryAR} simultaneously generates 4 or 16 tokens while maintaining high-quality image outputs.}
    \label{fig:decoding_case}
    \vspace{-8pt}
\end{figure}

The qualitative results presented in Figure~\ref{fig:decoding_case} further confirm QueryAR's exceptional robustness and coherence, generating high-quality images even at an aggressive $16\times$ acceleration ratio. This visual evidence corroborates our quantitative findings, confirming that critical visual details and semantic consistency are preserved despite the highly parallelized decoding. Collectively, these results confirm that QueryAR, through its innovative position-query mechanism, successfully learns a more stable and efficient autoregressive latent transition, effectively balancing superior generative performance with enhanced inference efficiency.

\section{Conclusion} \label{sec:conclusion}

In this work, we unlock the latent generative potential within large vision-language models by introducing {\methodname}, a visual generation tuning paradigm. This approach effectively activates generative capabilities without requiring architectural redesign or incurring prohibitive training costs. Our core innovation, {\methodname}-AE, meticulously aligns pretrained VLM semantic encoders with lightweight pixel decoders. This alignment directly addresses the inherent mismatch between traditional VAEs and autoregressive modeling, facilitating stable and efficient learning within a continuous latent space. Extensive experiments demonstrate that {\methodname}-AE achieves state-of-the-art reconstruction performance even under high compression ratios, and significantly advances autoregressive visual generation. {\methodname} delivers competitive scores on both GenEval and DPG-Bench, all while maintaining fast convergence and strong scalability. Beyond its specific model instance, {\methodname} provides a general blueprint for equipping any multimodal understanding model with high-quality visual generation capabilities. This work paves a promising path toward next-generation unified multimodal foundation models, capable of seamless perception-generation synergy, with future directions potentially extending the VGT framework in line with recent multimodal developments \cite{boow,varedit2025,liu2025omni,omnigen,x-omni}.

\section{Acknowledgements}

This work was supported by Kuaishou Technology. We extend our sincere gratitude to all collaborators involved in this project.

{
    \small
    \bibliographystyle{ieeenat_fullname}
    \bibliography{main}

@String(CVPR= {IEEE Conf. Comput. Vis. Pattern Recog.})

@String(ICLR = {Int. Conf. Learn. Represent.})

@String(CVPR  = {CVPR})

@String(ICLR  = {ICLR})

@article{vqvae,
  title={Neural discrete representation learning},
  author={Van Den Oord, Aaron and Vinyals, Oriol and others},
  journal={Advances in neural information processing systems},
  volume={30},
  year={2017}
}

@inproceedings{vqgan,
  title={Taming transformers for high-resolution image synthesis},
  author={Esser, Patrick and Rombach, Robin and Ommer, Bjorn},
  booktitle={Proceedings of the IEEE/CVF conference on computer vision and pattern recognition},
  pages={12873--12883},
  year={2021}
}

@inproceedings{simvq,
  title={Addressing representation collapse in vector quantized models with one linear layer},
  author={Zhu, Yongxin and Li, Bocheng and Xin, Yifei and Xia, Zhihua and Xu, Linli},
  booktitle={Proceedings of the IEEE/CVF International Conference on Computer Vision},
  pages={22968--22977},
  year={2025}
}

@inproceedings{cc12m,
  title={Conceptual 12m: Pushing web-scale image-text pre-training to recognize long-tail visual concepts},
  author={Changpinyo, Soravit and Sharma, Piyush and Ding, Nan and Soricut, Radu},
  booktitle={Proceedings of the IEEE/CVF conference on computer vision and pattern recognition},
  pages={3558--3568},
  year={2021}
}

@misc{echo4o,
      title={Echo-4o: Harnessing the Power of GPT-4o Synthetic Images for Improved Image Generation}, 
      author={Junyan Ye and Dongzhi Jiang and Zihao Wang and Leqi Zhu and Zhenghao Hu and Zilong Huang and Jun He and Zhiyuan Yan and Jinghua Yu and Hongsheng Li and Conghui He and Weijia Li},
      year={2025},
      eprint={2508.09987},
      archivePrefix={arXiv},
      primaryClass={cs.CV},
      url={https://arxiv.org/abs/2508.09987}, 
}

@misc{
sharegpt_4o,
author       = {Erfei Cui and Yinan He and Zheng Ma and Zhe Chen and Hao Tian and Weiyun Wang and Kunchang Li and Yi Wang and Wenhai Wang and Xizhou Zhu and Lewei Lu and Tong Lu and Yali Wang and Limin Wang and Yu Qiao and Jifeng Dai},
title        = {ShareGPT-4o: Comprehensive Multimodal Annotations With GPT-4o},
year         = {2024},
url          = {https://sharegpt4o.github.io/}, 
}

@article{journeydb,
  title={Journeydb: A benchmark for generative image understanding},
  author={Sun, Keqiang and Pan, Junting and Ge, Yuying and Li, Hao and Duan, Haodong and Wu, Xiaoshi and Zhang, Renrui and Zhou, Aojun and Qin, Zipeng and Wang, Yi and others},
  journal={Advances in neural information processing systems},
  volume={36},
  pages={49659--49678},
  year={2023}
}

@article{geneval,
  title={Geneval: An object-focused framework for evaluating text-to-image alignment},
  author={Ghosh, Dhruba and Hajishirzi, Hannaneh and Schmidt, Ludwig},
  journal={Advances in Neural Information Processing Systems},
  volume={36},
  pages={52132--52152},
  year={2023}
}

@article{dpgbench,
  title={Ella: Equip diffusion models with llm for enhanced semantic alignment},
  author={Hu, Xiwei and Wang, Rui and Fang, Yixiao and Fu, Bin and Cheng, Pei and Yu, Gang},
  journal={arXiv preprint arXiv:2403.05135},
  year={2024}
}

@inproceedings{ibq,
  title={Scalable image tokenization with index backpropagation quantization},
  author={Shi, Fengyuan and Luo, Zhuoyan and Ge, Yixiao and Yang, Yujiu and Shan, Ying and Wang, Limin},
  booktitle={Proceedings of the IEEE/CVF International Conference on Computer Vision},
  pages={16037--16046},
  year={2025}
}

@article{vqgan-lc,
  title={Scaling the codebook size of VQ-GAN to 100,000 with a utilization rate of 99\%},
  author={Zhu, Lei and Wei, Fangyun and Lu, Yanye and Chen, Dong},
  journal={Advances in Neural Information Processing Systems},
  volume={37},
  pages={12612--12635},
  year={2024}
}

@article{llamagen,
  title={Autoregressive model beats diffusion: Llama for scalable image generation},
  author={Sun, Peize and Jiang, Yi and Chen, Shoufa and Zhang, Shilong and Peng, Bingyue and Luo, Ping and Yuan, Zehuan},
  journal={arXiv preprint arXiv:2406.06525},
  year={2024}
}

@article{var,
  title={Visual autoregressive modeling: Scalable image generation via next-scale prediction},
  author={Tian, Keyu and Jiang, Yi and Yuan, Zehuan and Peng, Bingyue and Wang, Liwei},
  journal={Advances in neural information processing systems},
  volume={37},
  pages={84839--84865},
  year={2024}
}

@inproceedings{maskgit,
  title={Maskgit: Masked generative image transformer},
  author={Chang, Huiwen and Zhang, Han and Jiang, Lu and Liu, Ce and Freeman, William T},
  booktitle={Proceedings of the IEEE/CVF conference on computer vision and pattern recognition},
  pages={11315--11325},
  year={2022}
}

@article{emu3,
  title={Emu3: Next-token prediction is all you need},
  author={Wang, Xinlong and Zhang, Xiaosong and Luo, Zhengxiong and Sun, Quan and Cui, Yufeng and Wang, Jinsheng and Zhang, Fan and Wang, Yueze and Li, Zhen and Yu, Qiying and others},
  journal={arXiv preprint arXiv:2409.18869},
  year={2024}
}

@article{x-omni,
  title={X-omni: Reinforcement learning makes discrete autoregressive image generative models great again},
  author={Geng, Zigang and Wang, Yibing and Ma, Yeyao and Li, Chen and Rao, Yongming and Gu, Shuyang and Zhong, Zhao and Lu, Qinglin and Hu, Han and Zhang, Xiaosong and others},
  journal={arXiv preprint arXiv:2507.22058},
  year={2025}
}

@article{janus-pro,
  title={Janus-Pro: Unified Multimodal Understanding and Generation with Data and Model Scaling},
  author={Chen, Xiaokang and Wu, Zhiyu and Liu, Xingchao and Pan, Zizheng and Liu, Wen and Xie, Zhenda and Yu, Xingkai and Ruan, Chong},
  journal={arXiv preprint arXiv:2501.17811},
  year={2025}
}

@article{janus,
  title={Janus: Decoupling visual encoding for unified multimodal understanding and generation},
  author={Wu, Chengyue and Chen, Xiaokang and Wu, Zhiyu and Ma, Yiyang and Liu, Xingchao and Pan, Zizheng and Liu, Wen and Xie, Zhenda and Yu, Xingkai and Ruan, Chong and others},
  journal={arXiv preprint arXiv:2410.13848},
  year={2024}
}

@inproceedings{tokenflow,
  title={Tokenflow: Unified image tokenizer for multimodal understanding and generation},
  author={Qu, Liao and Zhang, Huichao and Liu, Yiheng and Wang, Xu and Jiang, Yi and Gao, Yiming and Ye, Hu and Du, Daniel K and Yuan, Zehuan and Wu, Xinglong},
  booktitle={Proceedings of the Computer Vision and Pattern Recognition Conference},
  pages={2545--2555},
  year={2025}
}

@article{blip3-o,
  title={Blip3-o: A family of fully open unified multimodal models-architecture, training and dataset},
  author={Chen, Jiuhai and Xu, Zhiyang and Pan, Xichen and Hu, Yushi and Qin, Can and Goldstein, Tom and Huang, Lifu and Zhou, Tianyi and Xie, Saining and Savarese, Silvio and others},
  journal={arXiv preprint arXiv:2505.09568},
  year={2025}
}

@inproceedings{clip,
  title={Learning transferable visual models from natural language supervision},
  author={Radford, Alec and Kim, Jong Wook and Hallacy, Chris and Ramesh, Aditya and Goh, Gabriel and Agarwal, Sandhini and Sastry, Girish and Askell, Amanda and Mishkin, Pamela and Clark, Jack and others},
  booktitle={International conference on machine learning},
  pages={8748--8763},
  year={2021},
  organization={PmLR}
}

@article{Ming-UniVision,
  title={Ming-UniVision: Joint Image Understanding and Generation with a Unified Continuous Tokenizer},
  author={Huang, Ziyuan and Zheng, DanDan and Zou, Cheng and Liu, Rui and Wang, Xiaolong and Ji, Kaixiang and Chai, Weilong and Sun, Jianxin and Wang, Libin and Lv, Yongjie and others},
  journal={arXiv preprint arXiv:2510.06590},
  year={2025}
}

@inproceedings{lpips,
  title={The unreasonable effectiveness of deep features as a perceptual metric},
  author={Zhang, Richard and Isola, Phillip and Efros, Alexei A and Shechtman, Eli and Wang, Oliver},
  booktitle={Proceedings of the IEEE conference on computer vision and pattern recognition},
  pages={586--595},
  year={2018}
}

@article{nextstep_1,
  title={Nextstep-1: Toward autoregressive image generation with continuous tokens at scale},
  author={Team, NextStep and Han, Chunrui and Li, Guopeng and Wu, Jingwei and Sun, Quan and Cai, Yan and Peng, Yuang and Ge, Zheng and Zhou, Deyu and Tang, Haomiao and others},
  journal={arXiv preprint arXiv:2508.10711},
  year={2025}
}

@inproceedings{fluid,
  author       = {Lijie Fan and
                  Tianhong Li and
                  Siyang Qin and
                  Yuanzhen Li and
                  Chen Sun and
                  Michael Rubinstein and
                  Deqing Sun and
                  Kaiming He and
                  Yonglong Tian},
  title        = {Fluid: Scaling Autoregressive Text-to-image Generative Models with
                  Continuous Tokens},
  booktitle    = {The Thirteenth International Conference on Learning Representations,
                  {ICLR} 2025, Singapore, April 24-28, 2025},
  publisher    = {OpenReview.net},
  year         = {2025},
}

@misc{fluxvae,
    author={Black Forest Labs},
    title={FLUX},
    year={2024},
    howpublished={\url{https://github.com/black-forest-labs/flux}},
}

@article{unilip,
  title={Unilip: Adapting clip for unified multimodal understanding, generation and editing},
  author={Tang, Hao and Xie, Chenwei and Bao, Xiaoyi and Weng, Tingyu and Li, Pandeng and Zheng, Yun and Wang, Liwei},
  journal={arXiv preprint arXiv:2507.23278},
  year={2025}
}

@article{mar,
  title={Autoregressive image generation without vector quantization},
  author={Li, Tianhong and Tian, Yonglong and Li, He and Deng, Mingyang and He, Kaiming},
  journal={Advances in Neural Information Processing Systems},
  volume={37},
  pages={56424--56445},
  year={2024}
}

@InProceedings{sd,
    author    = {Rombach, Robin and Blattmann, Andreas and Lorenz, Dominik and Esser, Patrick and Ommer, Bj\"orn},
    title     = {High-Resolution Image Synthesis With Latent Diffusion Models},
    booktitle = {Proceedings of the IEEE/CVF Conference on Computer Vision and Pattern Recognition (CVPR)},
    month     = {June},
    year      = {2022},
    pages     = {10684-10695}
}

@article{chameleo,
  author = {Chameleon Team},
  doi = {10.48550/arXiv.2405.09818},
  journal = {arXiv preprint arXiv:2405.09818},
  title = {Chameleon: Mixed-Modal Early-Fusion Foundation Models},
  url = {https://github.com/facebookresearch/chameleon},
  year = {2024}
}

@misc{sdxl,
      title={SDXL: Improving Latent Diffusion Models for High-Resolution Image Synthesis}, 
      author={Dustin Podell and Zion English and Kyle Lacey and Andreas Blattmann and Tim Dockhorn and Jonas Müller and Joe Penna and Robin Rombach},
      year={2023},
      eprint={2307.01952},
      archivePrefix={arXiv},
      primaryClass={cs.CV},
      url={https://arxiv.org/abs/2307.01952}, 
}

@misc{pixartalpha,
      title={PixArt-$\alpha$: Fast Training of Diffusion Transformer for Photorealistic Text-to-Image Synthesis}, 
      author={Junsong Chen and Jincheng Yu and Chongjian Ge and Lewei Yao and Enze Xie and Yue Wu and Zhongdao Wang and James Kwok and Ping Luo and Huchuan Lu and Zhenguo Li},
      year={2023},
      eprint={2310.00426},
      archivePrefix={arXiv},
      primaryClass={cs.CV}
}

@article{simplear,
  title={Simplear: Pushing the frontier of autoregressive visual generation through pretraining, sft, and rl},
  author={Wang, Junke and Tian, Zhi and Wang, Xun and Zhang, Xinyu and Huang, Weilin and Wu, Zuxuan and Jiang, Yu-Gang},
  journal={arXiv preprint arXiv:2504.11455},
  year={2025}
}

@article{Unitok,
  title={Unitok: A unified tokenizer for visual generation and understanding},
  author={Ma, Chuofan and Jiang, Yi and Wu, Junfeng and Yang, Jihan and Yu, Xin and Yuan, Zehuan and Peng, Bingyue and Qi, Xiaojuan},
  journal={arXiv preprint arXiv:2502.20321},
  year={2025}
}

@misc{gpt4,
  title = {GPT-4 Technical Report},
  author={Open, AI},
  year={2023}
}

@article{dalle2,
  author    = {Aditya Ramesh and Prafulla Dhariwal and Alex Nichol and Casey Chu and Mark Chen},
  title     = {Hierarchical Text-Conditional Image Generation with {CLIP} Latents},
  journal   = {arXiv preprint},
  volume    = {abs/2204.06125},
  year      = {2022},
  url       = {https://arxiv.org/abs/2204.06125},
  doi       = {10.48550/arXiv.2204.06125}
}

@article{gemini,
  title={Experiment with gemini 2.0 flash native image generation, 2025},
  author={Kampf, Kat and Brichtova, Nicole},
  journal={URL https://developers. googleblog. com/en/experiment-with-gemini-20-flash-native-image-generation},
  volume={3},
  year={2025}
}

@article{Qwen2_5_VL,
  title={Qwen2.5-VL Technical Report},
  author={Bai, Shuai and Chen, Keqin and Liu, Xuejing and Wang, Jialin and Ge, Wenbin and Song, Sibo and Dang, Kai and Wang, Peng and Wang, Shijie and Tang, Jun and Zhong, Humen and Zhu, Yuanzhi and Yang, Mingkun and Li, Zhaohai and Wan, Jianqiang and Wang, Pengfei and Ding, Wei and Fu, Zheren and Xu, Yiheng and Ye, Jiabo and Zhang, Xi and Xie, Tianbao and Cheng, Zesen and Zhang, Hang and Yang, Zhibo and Xu, Haiyang and Lin, Junyang},
  journal={arXiv preprint arXiv:2502.13923},
  year={2025}
}

@article{internvl3,
  title={Internvl3: Exploring advanced training and test-time recipes for open-source multimodal models},
  author={Zhu, Jinguo and Wang, Weiyun and Chen, Zhe and Liu, Zhaoyang and Ye, Shenglong and Gu, Lixin and Tian, Hao and Duan, Yuchen and Su, Weijie and Shao, Jie and others},
  journal={arXiv preprint arXiv:2504.10479},
  year={2025}
}

@article{intern-s1,
  title={Intern-s1: A scientific multimodal foundation model},
  author={Bai, Lei and Cai, Zhongrui and Cao, Yuhang and Cao, Maosong and Cao, Weihan and Chen, Chiyu and Chen, Haojiong and Chen, Kai and Chen, Pengcheng and Chen, Ying and others},
  journal={arXiv preprint arXiv:2508.15763},
  year={2025}
}

@article{Dualtoken,
  title={Dualtoken: Towards unifying visual understanding and generation with dual visual vocabularies},
  author={Song, Wei and Wang, Yuran and Song, Zijia and Li, Yadong and Sun, Haoze and Chen, Weipeng and Zhou, Zenan and Xu, Jianhua and Wang, Jiaqi and Yu, Kaicheng},
  journal={arXiv preprint arXiv:2503.14324},
  year={2025}
}

@article{Vila-u,
  title={Vila-u: a unified foundation model integrating visual understanding and generation},
  author={Wu, Yecheng and Zhang, Zhuoyang and Chen, Junyu and Tang, Haotian and Li, Dacheng and Fang, Yunhao and Zhu, Ligeng and Xie, Enze and Yin, Hongxu and Yi, Li and others},
  journal={arXiv preprint arXiv:2409.04429},
  year={2024}
}

@article{rae,
  title={Diffusion Transformers with Representation Autoencoders},
  author={Zheng, Boyang and Ma, Nanye and Tong, Shengbang and Xie, Saining},
  journal={arXiv preprint arXiv:2510.11690},
  year={2025}
}

@article{dcae,
 title={Deep Compression Autoencoder for Efficient High-Resolution Diffusion Models},
 author={Chen, Junyu and Cai, Han and Chen, Junsong and Xie, Enze and Yang, Shang and Tang, Haotian and Li, Muyang and Lu, Yao and Han, Song},
 journal={arXiv preprint arXiv:2410.10733},
 year={2024}
}

@article{svg,
      title={Latent Diffusion Model without Variational Autoencoder}, 
      author={Minglei Shi and Haolin Wang and Wenzhao Zheng and Ziyang Yuan and Xiaoshi Wu and Xintao Wang and Pengfei Wan and Jie Zhou and Jiwen Lu},
      year={2025},
      journal={arXiv preprint arXiv:2510.15301},
}

@article{spherear,
      title={Hyperspherical Latents Improve Continuous-Token Autoregressive Generation}, 
      author={Guolin Ke and Hui Xue},
      year={2025},
    journal={arXiv preprint arXiv:2509.24335},
}

@inproceedings{repa,
  title={Representation Alignment for Generation: Training Diffusion Transformers Is Easier Than You Think},
  author={Sihyun Yu and Sangkyung Kwak and Huiwon Jang and Jongheon Jeong and Jonathan Huang and Jinwoo Shin and Saining Xie},
  year={2025},
  booktitle={International Conference on Learning Representations},
}

@article{repae,
  title={REPA-E: Unlocking VAE for End-to-End Tuning with Latent Diffusion Transformers},
  author={Xingjian Leng and Jaskirat Singh and Yunzhong Hou and Zhenchang Xing and Saining Xie and Liang Zheng},
  year={2025},
  journal={arXiv preprint arXiv:2504.10483},
}

@article{vae,
  title={Auto-encoding variational bayes},
  author={Kingma, Diederik P and Welling, Max},
  journal={arXiv preprint arXiv:1312.6114},
  year={2013}
}

@inproceedings{kl-16,
    author    = {Rombach, Robin and Blattmann, Andreas and Lorenz, Dominik and Esser, Patrick and Ommer, Bj\"orn},
    title     = {High-Resolution Image Synthesis With Latent Diffusion Models},
    booktitle = {Proceedings of the IEEE/CVF Conference on Computer Vision and Pattern Recognition (CVPR)},
    month     = {June},
    year      = {2022},
    pages     = {10684-10695}
}

@article{sun2024multimodal,
  title={Multimodal latent language modeling with next-token diffusion},
  author={Sun, Yutao and Bao, Hangbo and Wang, Wenhui and Peng, Zhiliang and Dong, Li and Huang, Shaohan and Wang, Jianyong and Wei, Furu},
  journal={arXiv preprint arXiv:2412.08635},
  year={2024}
}

@article{seedx,
  title={SEED-X: Multimodal Models with Unified Multi-granularity Comprehension and Generation},
  author={Ge, Yuying and Zhao, Sijie and Zhu, Jinguo and Ge, Yixiao and Yi, Kun and Song, Lin and Li, Chen and Ding, Xiaohan and Shan, Ying},
  journal={arXiv preprint arXiv:2404.14396},
  year={2024}
}

@inproceedings{noisetimestep,
  author       = {Patrick Esser and
                  Sumith Kulal and
                  Andreas Blattmann and
                  Rahim Entezari and
                  Jonas M{\"{u}}ller and
                  Harry Saini and
                  Yam Levi and
                  Dominik Lorenz and
                  Axel Sauer and
                  Frederic Boesel and
                  Dustin Podell and
                  Tim Dockhorn and
                  Zion English and
                  Robin Rombach},
  title        = {Scaling Rectified Flow Transformers for High-Resolution Image Synthesis},
  booktitle    = {Forty-first International Conference on Machine Learning, {ICML} 2024,
                  Vienna, Austria, July 21-27, 2024},
  publisher    = {OpenReview.net},
  year         = {2024},
}

@article{randar,
    title={RandAR: Decoder-only Autoregressive Visual Generation in Random Orders},
    author={Pang, Ziqi and Zhang, Tianyuan and Luan, Fujun and Man, Yunze and Tan, Hao and Zhang, Kai and Freeman, William T. and Wang, Yu-Xiong},
    journal={arXiv preprint arXiv:2412.01827},
    year={2024}
}

@article{rar,
  author    = {Qihang Yu and Ju He and Xueqing Deng and Xiaohui Shen and Liang-Chieh Chen},
  title     = {Randomized Autoregressive Visual Generation},
  journal   = {arxiv},
  year      = {2024}
}

@inproceedings{givt,
  title={Givt: Generative infinite-vocabulary transformers},
  author={Tschannen, Michael and Eastwood, Cian and Mentzer, Fabian},
  booktitle={European Conference on Computer Vision},
  pages={292--309},
  year={2024},
  organization={Springer}
}

@inproceedings{dino,
  title={Emerging properties in self-supervised vision transformers},
  author={Caron, Mathilde and Touvron, Hugo and Misra, Ishan and J{\'e}gou, Herv{\'e} and Mairal, Julien and Bojanowski, Piotr and Joulin, Armand},
  booktitle={Proceedings of the IEEE/CVF international conference on computer vision},
  pages={9650--9660},
  year={2021}
}

@article{clipgen,
  title={CLIP-GEN: Language-Free Training of a Text-to-Image Generator with CLIP},
  author={Wang, Zihao and Liu, Wei and He, Qian and Wu, Xinglong and Yi, Zili},
  journal={arXiv preprint arXiv:2203.00386},
  year={2022}
}

@article{vqkd,
  title={Beit v2: Masked image modeling with vector-quantized visual tokenizers},
  author={Peng, Zhiliang and Dong, Li and Bao, Hangbo and Ye, Qixiang and Wei, Furu},
  journal={arXiv preprint arXiv:2208.06366},
  year={2022}
}

@article{mme,
  title={MME: A Comprehensive Evaluation Benchmark for Multimodal Large Language Models},
  author={Fu, Chaoyou and Chen, Peixian and Shen, Yunhang and Qin, Yulei and Zhang, Mengdan and Lin, Xu and Yang, Jinrui and Zheng, Xiawu and Li, Ke and Sun, Xing and others},
  journal={arXiv preprint arXiv:2306.13394},
  year={2023}
}

@inproceedings{textvqa,
  title={Towards vqa models that can read},
  author={Singh, Amanpreet and Natarajan, Vivek and Shah, Meet and Jiang, Yu and Chen, Xinlei and Batra, Dhruv and Parikh, Devi and Rohrbach, Marcus},
  booktitle={Proceedings of the IEEE/CVF conference on computer vision and pattern recognition},
  pages={8317--8326},
  year={2019}
}

@inproceedings{mmbench,
  title={Mmbench: Is your multi-modal model an all-around player?},
  author={Liu, Yuan and Duan, Haodong and Zhang, Yuanhan and Li, Bo and Zhang, Songyang and Zhao, Wangbo and Yuan, Yike and Wang, Jiaqi and He, Conghui and Liu, Ziwei and others},
  booktitle={European conference on computer vision},
  pages={216--233},
  year={2024},
  organization={Springer}
}

@misc{ai2d, 
    title={A Diagram Is Worth A Dozen Images}, 
    author={Aniruddha Kembhavi and Mike Salvato and Eric Kolve and Minjoon Seo and Hannaneh Hajishirzi and Ali Farhadi}, 
    year={2016}, 
    eprint={1603.07396}, 
    archivePrefix={arXiv}, 
    primaryClass={cs.CV} 
}

@article{open-magvit2,
  title={Open-magvit2: An open-source project toward democratizing auto-regressive visual generation},
  author={Luo, Zhuoyan and Shi, Fengyuan and Ge, Yixiao and Yang, Yujiu and Wang, Limin and Shan, Ying},
  journal={arXiv preprint arXiv:2409.04410},
  year={2024}
}

@inproceedings{vavae,
  title={Reconstruction vs. generation: Taming optimization dilemma in latent diffusion models},
  author={Yao, Jingfeng and Yang, Bin and Wang, Xinggang},
  booktitle={Proceedings of the IEEE/CVF Conference on Computer Vision and Pattern Recognition},
  year={2025}
}

@article{vavae_dit,
  title={Fasterdit: Towards faster diffusion transformers training without architecture modification},
  author={Yao, Jingfeng and Wang, Cheng and Liu, Wenyu and Wang, Xinggang},
  journal={Advances in Neural Information Processing Systems},
  volume={37},
  pages={56166--56189},
  year={2024}
}

@inproceedings{boow,
  author       = {Xuanpu Zhang and
                  Dan Song and
                  Pengxin Zhan and
                  Tianyu Chang and
                  Jianhao Zeng and
                  Qingguo Chen and
                  Weihua Luo and
                  An{-}An Liu},
  title        = {BooW-VTON: Boosting In-the-Wild Virtual Try-On via Mask-Free Pseudo
                  Data Training},
  booktitle    = {{IEEE/CVF} Conference on Computer Vision and Pattern Recognition,
                  {CVPR} 2025, Nashville, TN, USA, June 11-15, 2025},
  pages        = {26399--26408},
  publisher    = {Computer Vision Foundation / {IEEE}},
  year         = {2025}
}

@article{liu2025omni,
  title={Omni-Dish: Photorealistic and Faithful Image Generation and Editing for Arbitrary Chinese Dishes},
  author={Liu, Huijie and Wang, Bingcan and Hu, Jie and Wei, Xiaoming and Kang, Guoliang},
  journal={arXiv preprint arXiv:2504.09948},
  year={2025}
}

@article{varedit2025,
  title={Visual Autoregressive Modeling for Instruction-Guided Image Editing},
  author={Mao, Qingyang and Cai, Qi and Li, Yehao and Pan, Yingwei and Cheng, Mingyue and Yao, Ting and Liu, Qi and Mei, Tao},
  journal={arXiv preprint arXiv:2508.15772},
  year={2025}
}

@article{omnigen,
  title={Omnigen: Unified image generation},
  author={Xiao, Shitao and Wang, Yueze and Zhou, Junjie and Yuan, Huaying and Xing, Xingrun and Yan, Ruiran and Wang, Shuting and Huang, Tiejun and Liu, Zheng},
  journal={arXiv preprint arXiv:2409.11340},
  year={2024}
}
}









\end{document}